\documentclass[a4paper,fleqn]{cas-sc}
\ExplSyntaxOn
\keys_define:nn { cas / fig }
  { unknown .code:n = \tl_set:Nx \l_fig_pos_tl { \l_keys_key_str } }
\ExplSyntaxOff
\ExplSyntaxOn
\keys_define:nn { cas / tbl }
  {
    unknown .code:n =
      \str_if_eq:VnTF \l_keys_key_str { H }
        { \tl_set:Nn \l_tbl_pos_tl { !htbp } }
        { \tl_set:Nx \l_tbl_pos_tl { \l_keys_key_str } }
  }
\ExplSyntaxOff
\usepackage{bm}
\usepackage{xspace}
\usepackage{float}
\usepackage{caption}
\usepackage{subcaption}
\usepackage[numbers,sort&compress]{natbib}
\usepackage{dirtytalk}
\usepackage[utf8]{inputenc}
\usepackage{siunitx}
\usepackage{tikz}
\usetikzlibrary{arrows.meta,positioning,calc,backgrounds}
\definecolor{mypink}{HTML}{EF7B6B}
\definecolor{myblue}{HTML}{35A2D4}
\newcommand{\ff}{FAST.Farm\xspace}
\newcommand{\floris}{FLORIS\xspace}
\newcommand{\ms}{~\si{\meter\per\second}\xspace}

\begin{document}
\shorttitle{Noise Steering}
\shortauthors{M. de Frutos et al.}
\title[mode=title]{Noise Steering: Exploiting Wake Steering and Acoustic Directivity for Noise-Constrained Wind Farm Control via Reinforcement Learning}

\author[inst1]{Martín de\,Frutos}[orcid=0009-0005-3357-630X]
\cormark[1]
\fnmark[1]
\ead{m.defrutos@upm.es}
\credit{Conceptualization, Investigation, Methodology, Software, Validation, Visualization, Writing -- original draft}

\author[inst2,inst5,inst3]{Baptiste Corban}[orcid=0009-0008-3412-6660]
\fnmark[1]
\credit{Conceptualization, Investigation, Methodology, Software, Validation, Visualization, Writing -- original draft}

\author[inst6]{Claire Bizon Monroc}[orcid=0000-0002-7614-5359]
\credit{Methodology, Software, Writing -- review \& editing}

\author[inst2,inst5]{Ana Bušić}[orcid=0000-0002-4133-3739]
\credit{Conceptualization, Methodology,  Supervision, Validation, Project administration, Writing -- review \& editing}

\author[inst3]{Donatien Dubuc}[orcid=0000-0002-8300-1353]
\credit{Supervision, Project administration, Writing -- review \& editing}

\author[inst3]{Jiamin Zhu}[orcid=0000-0002-4552-5519]
\credit{Supervision, Project administration, Writing -- review \& editing}

\author[inst1,inst4]{Esteban Ferrer}[orcid=0000-0003-1519-0444]
\credit{Conceptualization, Methodology, Supervision, Validation, Project administration, Funding acquisition, Writing -- review \& editing}

\affiliation[inst1]{organization={ETSIAE-UPM - School of Aeronautics, Universidad Politécnica de Madrid},
            addressline={Plaza Cardenal Cisneros 3},
            city={Madrid},
            postcode={E-28040},
            country={Spain}}

\affiliation[inst2]{organization={Inria},
            addressline={48 rue Barrault},
            city={Paris},
            postcode={75013},
            country={France}}

\affiliation[inst5]{organization={DI ENS, Ecole Normale Supérieure, PSL Research University},
            city={Paris},
            country={France}}

\affiliation[inst3]{organization={IFP Energies nouvelles},
            addressline={1 et 4 avenue de Bois-Préau},
            city={Rueil-Malmaison},
            postcode={92852},
            country={France}}

\affiliation[inst4]{organization={Center for Computational Simulation, Universidad Politécnica de Madrid},
            addressline={Campus de Boadilla del Monte},
            city={Madrid},
            postcode={E-28660},
            country={Spain}}

\affiliation[inst6]{organization={Center for Processes, Renewable Energy and Energy Systems (PERSEE), Mines Paris PSL},
            addressline={1 Rue Claude Daunesse},
            city={Sophia Antipolis},
            postcode={06560},
            country={France}}

\cortext[1]{Corresponding author}
\fntext[1]{These authors contributed equally to this work.}
\maketitle
\begin{abstract}
Existing noise mitigation strategies for wind farms usually consist in reducing turbines power output, without accounting for the wake interactions between them and the noise emission directivity. This leads to over-conservative policies and unexploited wind power. 
Yet, turbine yaw can simultaneously affect wake interactions and the direction of noise emission, offering an additional degree of freedom for farm-level noise control. We therefore propose \textit{Noise Steering}, a model-free reinforcement learning (RL) strategy that jointly controls turbine yaw angles and power setpoints to maximise farm power while satisfying a local noise constraint in a defined region. To enable training at a reduced computational cost, a low-fidelity acoustic model is coupled with the steady-state \floris simulator. The RL agent effectively learns how to redirect wakes and noise directivity away from the protected region while redistributing power across the farm. On a six-turbine test farm, the policy satisfies the \SI{40}{\decibel} noise constraint throughout the full range of turbulent winds, with an average power deficit of only 2\% relative to \textit{greedy unimpeded control}, and up to 7.7\% at high wind speeds. In comparison, the industry-standard noise-reduced operation (NRO) baseline here generates an average power loss of 17\%. When deployed in the higher-fidelity FAST.Farm simulator, the policy largely retains its performance under steady and turbulent inflow, despite differences in simulation fidelity. These results demonstrate the potential of combining yaw control and power derating through RL to improve the power-noise trade-off in wind farm operation.
\end{abstract}

%% Keywords

% \newpage
\section{Introduction}
Wind energy has become one of the fastest-growing sources of renewable electricity generation,  with global installed capacity rising from \SI{417}{\giga\watt} in 2015 to more than
\SI{1130}{\giga\watt} by the end of 2024, which now accounts for about a quarter of all renewable power capacity worldwide~\citep{IRENA2025}. This growth, however, has brought turbines closer to populated areas, and the aerodynamic noise they emit is recognised as a source of public annoyance \citep{wolsink1993annoyance,pedersen2009response}. As a result, noise regulations have been introduced to limit the sound levels that wind farms may produce, often imposing significant penalties on operators that exceed the allowed thresholds \citep{DAVY2018288,alamir2021penalties}. 

Existing noise mitigation strategies mainly operate at the turbine level. Blade design optimisation \citep{bortolotti2018integrated, KOU2023119513} and other approaches, such as blade trailing edge serrations \citep{DESHMUKH2019633}, have been explored. On the turbine control side, industry standards define noise-reduced operation (NRO) modes for individual turbines \citep{van2025sound}, effectively limiting their rotational speed and power output.  \citep{ducExperimentalInvestigationEffect2026} provide an example of acoustic curtailment look-up table used to control the turbines. Model predictive control was used in \citep{shaltout2015tradeoff} to find the optimal turbine controls (blade pitch and generator torque) to satisfy a noise level constraint.
At the farm level, noise mitigation has mainly been addressed through layout optimisation \citep{cao2020wind,wu2020optimizing} and optimisation of individual NRO levels \citep{shah2026leveraging,nyborg2022optimization,sorensenProductionOptimizedNoise}. 
Moreover, \citet{shah2026leveraging} define uniform power derating of the turbines as a common control strategy by wind farm operators.
These methods enable compliance with the noise regulation but reduce power generation of the wind farm, which leads to reduced revenue in real operation. The median loss in annual energy production (AEP) was estimated to be of 2.8\% in France between 2020 and 2022 \citep{Willis2023Curtailment}.

Meanwhile, an important share of wind farm control studies focuses on wake steering for power maximisation of the wind farm \citep{boersmaTutorialControlorientedModeling2017}. By misaligning turbines, yaw control redirects their wakes and can reduce velocity deficits for downstream turbines. Yaw control can  also modify the spatial distribution (or directivity) of turbine noise. Wind turbine noise exhibits a dipole radiation pattern approximately aligned with the rotor axis (i.e., maximum with respect to the rotor normal) \citep{oerlemans2009prediction,okada2016horizontal}, such that yawing a turbine changes the direction of maximum noise emission.
The potential of exploiting noise directivity in wind farm operations has recently been highlighted by \citet{shah2026leveraging}. However, existing noise-reduction control strategies rely only on power adjustments and do not exploit turbine yaw control \citep{ducExperimentalInvestigationEffect2026}. 
In this article, we extend the wake steering strategy, which targets power maximisation, to a novel farm-level noise-reduced operation strategy, referred to as \textit{Noise Steering}. By controlling the yaw angles and NRO levels of the wind turbines, this control technique can maximise power generation while respecting a localised noise constraint.

Wind farm control has traditionally focused on model-based control to coordinate turbines within a farm \citep{meyersWFC}.
These approaches can similarly be applied to \textit{Noise Steering}, for example through Model Predictive Control or offline model-based optimisation.
However, these methods are limited by the precision of their underlying model and generally require solving an optimisation problem for each operating condition. Reinforcement Learning (RL) has recently emerged as a promising data-driven alternative \citep{Dong_2022,ABKAR2023100475}, with the potential to provide a model-free closed-loop controller whose policy is computationally inexpensive to evaluate once trained and that can continue to learn and adapt online when deployed on a wind farm.

It has recently shown promising results on coordinated wind farm control for power maximisation \citep{dong2021intelligent,moleReinforcementLearningIncreases2025,monrocActorCriticAgents2023, Ebrahimi_2026}.
However, noise constraints have not yet been considered in these learning-based wind farm control approaches. 

To address this gap, we propose a cooperative wind farm RL control to perform \textit{Noise Steering}. 
RL training requires a large number of simulations, making high-fidelity noise  evaluation (e.g. Large Eddy Simulation of farms, \citep{rismondo2026wind}) computationally prohibitive. Therefore, we extend the steady-state \floris simulator~\citep{gebraadWindPlantPower2016} with a low-fidelity acoustic model to efficiently estimate farm-level noise emissions. We combine the analytical model of \citet{hagg1990aerodynamic} with an enforced dipole directivity, and validate it against FAST.Farm BPM acoustic model \citep{brooks1989airfoil}.
The RL agent learns to maximise farm power production while respecting a noise constraint, namely a maximum sound level allowed at a protected receiver region near the farm. Unlike turbine-level NRO or static layout optimisation, our formulation allows the agent to combine both yaw misalignment and power derating for each turbine. This enables to increase power generation compared to standard strategies by reducing wake interactions and redirecting noise directivity patterns away from the protected region. 
The policy, trained exclusively under steady inflow in \floris, is then deployed without retraining in the higher-fidelity dynamic simulator FAST.Farm, both under steady and unsteady inflow. 

The remainder of this paper is organised as follows: \autoref{sec:Methodology} defines the control problem and wind farm case, as well as the RL set-up and noise estimation in \floris. Then, \autoref{sec:floris_results} introduces the training, evaluation and analysis of the RL \textit{Noise Steering} agent. Finally, \autoref{sec:FF_results} details the deployment in the FAST.Farm higher-fidelity dynamic simulator, first under steady inflow then under unsteady turbulent inflow. Conclusions and perspectives are given in \autoref{sec:conclusions}.

\section{Methodology}
\label{sec:Methodology}
\subsection{Wind farm control problem and set-up}

We study a wind farm and a control region where the sound level must remain below a given limit. This can represent nearby buildings and habitations for an onshore wind farm subject to noise regulations. The objective is to control the wind turbines to maximise the power output while keeping the maximum sound level in the control region below the limit.

\begin{figure}
    \centering
    \input{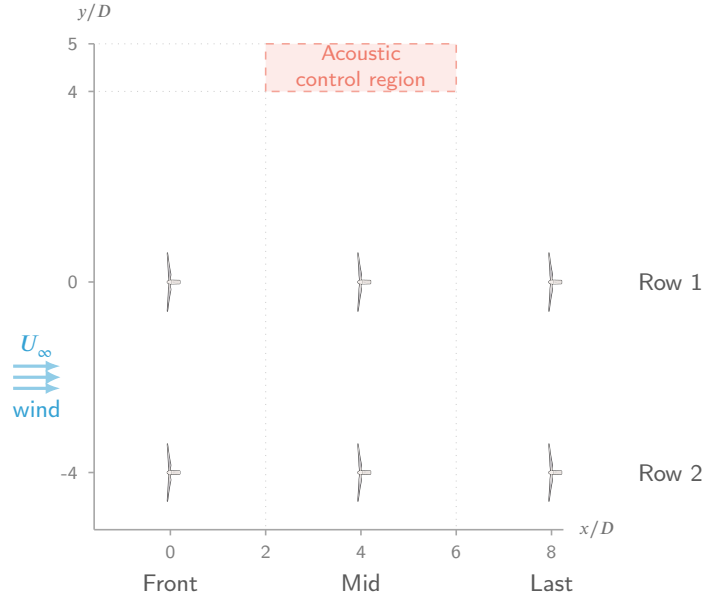}
    \caption{Wind farm layout and acoustic control region location.}
    \label{fig:wf_layout}
\end{figure}

The wind farm studied consists of six turbines, as shown in~\autoref{fig:wf_layout}. The turbines are arranged in two rows of three turbines, with a spacing of 4 diameters between consecutive turbines. Here, we use the NREL 5MW reference turbine~\citep{jonkmanDefinition5MWReference2009}, with a rotor diameter $D$ of \SI{126}{\meter}, but the methodology is general and can be adapted to other turbines. The control region is located $4D$ away from the top row and at the height of the ground. \citet{alamir2021penalties} reports significant variation in wind turbine noise regulations in different countries, with limits ranging from \SI{35}{\decibel} to \SI{60}{\decibel}. In this study, the noise limit is set to \SI{40}{\decibel}. This value is representative of European regulations~\citep{nieuwenhuizenDifferencesNoiseRegulations2015}, and chosen to impose a restrictive constraint, \SI{4}{dB} below the maximum sound level obtained for this wind farm in our simulations. Under normal operating conditions, the sound level in the control region exceeds this limit, except at low wind speeds. Note that the proposed methodology and the RL framework are not specific to the 40 dB limit and can be applied to other noise thresholds.

In this work, we consider different wind inflow velocities while keeping the wind direction constant and aligned with the two turbine rows. The small spacing between turbines and the aligned inflow result in strong wake interactions. 

To maximise farm power while satisfying a noise constraint, we consider the following control problem:

\begin{equation}
    \max_{\mathbf{u}} \; P_{\mathrm{farm}}(\mathbf{u})
    \quad \text{s.t.} \quad
    \max_{\bm x\in\mathcal{R}} L_p^F(\bm x,\mathbf{u}) \leq L_{p,\mathrm{thr}},
\label{eq:optim_problem}
\end{equation}
where $\mathbf{u}$ contains the power and yaw setpoints, $L_p^F$ denotes the acoustic emissions in \si{\decibel} of the wind farm at each location $\bm x$ and $L_{p,\mathrm{thr}}$ stands for the desired noise constraint. The control region is denoted by $\mathcal{R}$. 

Yaw setpoints enable wake steering, while power setpoints curtail the maximum power output of the wind turbine, controlling the turbine induction. The noise constraint is controlled by leveraging the effects of these controls on both the noise amplitude and its directivity. We refer to this control strategy as \textit{Noise Steering}. 
Wind farms are usually operated by maximising the individual powers of each turbine, i.e. by having each turbine facing the wind maximising its power output given the incoming wind. This is referred to as \textit{greedy control}.

The wind farm is modelled using two simulators, \floris~\citep{gebraadWindPlantPower2016} and \ff~\citep{jonkmanDevelopmentFASTFarmNew2017}, which have different levels of fidelity and computational costs.
\floris is a steady-state simulator, and the wind farm is simulated through a sequence of steady states, without resolving the transient propagation of the wakes. It does not include acoustic models, but provides fast results. \ff is a blade element momentum (BEM)-based solver that resolves the temporal dynamics of the wind farm, including wake propagation. It provides higher-fidelity modelling and is coupled with several multi-physics models, including acoustics. As described by \citet{bortolotti2020aeroacoustics}, the aeroacoustic modelling in \ff is based on the Brooks-Pope-Marcolini (BPM) semi-empirical noise model.
This higher fidelity comes at an increased computational cost. Both simulators are interfaced with Python using the WFCRL interface by \citet{wfcrl}.

\subsection{Reinforcement Learning for wind farm control}

We choose to formulate the wind farm control problem \eqref{eq:optim_problem} as a reinforcement learning problem. 
At each time step, the RL agent observes the state of the wind farm and selects power and yaw setpoints for each turbine. 
The state vector is defined as
\begin{equation}
    \mathbf s = [U_\infty, \mathbf P, \bm{\gamma_t}],
\end{equation}
where $U_\infty$ is the free-stream wind speed, $\mathbf P$ contains the current turbine powers, and $\bm{\gamma_t}$ contains the current turbine yaw angles. 

The action of the agent consists of increments in the power and yaw setpoints. The yaw actions are defined as
$\Delta\bm{\gamma_t}^i \in [\SI{-5}\degree,\SI5\degree]$,
with the yaw angles constrained to the range $\bm{\gamma_t}^i \in [\SI{-40}\degree,\SI{40}\degree]$ for each wind turbine.
Similarly, the power setpoints $\mathbf{P_S}$ are updated by increments $\Delta \mathbf{P_S} \in[-1,1]$~\si{\mega\watt}, with the resulting setpoints being constrained to [0,5]~\si{\mega\watt}.
In \floris, successive steady states are simulated; therefore, a new action is taken at each of these steps. In \ff, an action is taken every $\Delta t_\text{RL}=3$ seconds.

The objective of the RL agent is to maximise wind farm power while satisfying the noise constraint.
Therefore, the reward is defined as
\begin{equation}
\label{eq:reward_def}
r =
\frac{\sum_i P_i}{P_\mathrm{greedy}}
-
\underbrace{
    \alpha \mathrm{ReLU}\left(\hat{L}_p-L_{p,\mathrm{thr}}\right)
}_{\text{Noise threshold penalty}}
-
\underbrace{
    \vphantom{\alpha \mathrm{ReLU}\left(\hat{L}_p-L_{p,\mathrm{thr}}\right)}
    \beta_1\lVert\Delta\bm{\gamma_t}\rVert_2
    - \beta_2\lVert\Delta\mathbf{P_S}\rVert_2
    - \beta_3\lVert\mathbf P-\mathbf{P_S}\rVert_2
}_{\text{action penalties}}
\end{equation}
where $P_i$ is the power produced by the turbine $i$, $P_\mathrm{greedy}$ is the farm power obtained when the farm is operated with \textit{greedy control}.
$\hat{L}_p$ is the maximum sound level measured in the control region. The parameter $L_{p,\mathrm{thr}}$ is the noise limit (here \SI{40}{\decibel}). The first term therefore rewards power production, while the second penalises violations of the noise constraint. The last three terms penalise large changes in yaw angles, large changes in the power setpoints, and differences between the current powers and the power setpoints, respectively. These helped the training process, notably by reducing oscillations and avoiding unfeasible actions. If an action results in a yaw angle outside the allowed range, a reward of $r=-3$ is assigned.

We use Proximal Policy Optimisation (PPO) to train the RL agent. PPO is a state-of-the-art actor-critic method that updates the policy while limiting the change between successive policies, \citep{schulman2017proximalpolicyoptimizationalgorithms}. 
The algorithm hyperparameters, as well as the reward coefficients used are detailed in \autoref{app:hyperparameters}.

RL training requires a large number of interactions with the environment. Running these interactions with \ff would therefore result in a prohibitive computational cost. Therefore, we use \floris during training, which provides a much lower computational cost. The resulting policies are then evaluated using \ff to assess their performance in a higher-fidelity, time-resolved environment.

\subsection{\floris to \ff calibration}
\label{sec:calibration}

\citet{Ebrahimi_2026} highlight the differences in predicted wake losses between \floris and \ff, and propose a recalibration of the \floris models. It is also common in the industry to calibrate low-fidelity models with field measurements to improve accuracy.
We perform a similar calibration to reduce the model mismatch in power between the simulators and help the transfer from \floris to \ff.

The calibration is performed over a wind speed range of 8~\ms to 15~\ms, corresponding to the range used for training the RL agent. A dataset comprising 16 \ff simulations is generated to calibrate the parameters of the Gaussian wake model~\citep{bastankhahNewAnalyticalModel2014, niayifarAnalyticalModelingWind2016} implemented in \floris. The simulations are performed until a stationary wind flow is reached and covered different yaw conditions, consisting in aligned operation and yaw misalignments of 10\si{\degree}, 20\si{\degree}, and 30\si{\degree}, as well as different wind speed values. All simulations are conducted using the same wind farm layout shown in \autoref{fig:wf_layout}.

A differential evolution optimiser \citep{storn1997differential} is employed for calibration. The optimisation objective is to minimise the difference between the turbine power outputs predicted by the two solvers. The calibration strategy consisted in two phases. First, the wake deficit parameters are calibrated using only the aligned cases in the dataset. Afterwards, these parameters are frozen and the complete dataset is used to optimise the wake deflection parameters. The parameters obtained through this calibration process are described in \autoref{tab:wake_parameters}. 

\begin{table}[htbp]
    \centering
    \caption{Parameters of the Gaussian wake model used in \floris.}
    \label{tab:wake_parameters}
    \begin{tabular}{lcccccc}
        \toprule
        & \multicolumn{4}{c}{Velocity deficit} &
        \multicolumn{2}{c}{Deflection} \\
        \cmidrule(lr){2-5}
        \cmidrule(l){6-7}
        & $k_a$ & $k_b$ & $\alpha_g$ & $\beta_g$ & $d_m$ & $b_d$ \\
        \midrule
        \floris default & 0.38 & 0.004 & 0.58 &  0.077 & 1.0 & 0.0 \\
        \textbf{Calibrated} & 0.31 & 0.0024 & 0.99 & 0.149 & 0.893 & 0.00699 \\
        \bottomrule
    \end{tabular}
\end{table}

The initial modelling mismatch between \floris and \ff for the computed data set gives a mean absolute percentage error (MAPE) of 15\% (7\% for aligned cases). After calibration, this MAPE decreases to 2\% for the aligned cases and to 7\% for the complete data set. Overall, the simulation gap between \ff and \floris is strongly reduced by the calibration. 

\subsection{Aeroacoustic modelling}
\label{sec:aeroacoustic}
The acoustic impact of the wind farm is assessed using noise prediction models, as described in \citep{wagner2012wind}. There are different approaches to prediction of wind turbine noise. \ff has a semi-empirical method implemented based on the BPM model, \citep{brooks1989airfoil}. This would be a Class II noise prediction code according to the classification proposed by \citet{lowsonnoise}. This class of methods is generally considered to be a mid-fidelity approach capable of capturing the main noise-generation mechanisms of wind turbines. These methods use the output of the BEM codes to account for the influence of turbine operating conditions on the predicted noise levels. 

In contrast, Class I methods within the same classification are lower-fidelity analytical approaches, generally based on empirical relationships. \floris does not include a wind turbine noise prediction model. As a low-fidelity wind turbine simulator, it does not provide the in-blade information required by Class II semi-empirical noise prediction methods. Therefore, a Class I model is implemented for noise prediction in \floris.
The A-weighted overall sound pressure level (OASPL or $L_{p,A}$) of a single wind turbine is estimated by combining the model proposed by \citet{hagg1990aerodynamic} with an enforced directivity function. The resulting analytical model for the OASPL is given by
\begin{equation}
    \label{eq:noise_model}
    %\begin{aligned} 
    L_{p,A}(\bm x)=C_{1}\mathrm{log}_{10}V_{\mathrm{tip}}+C_{2}\mathrm{log}_{10}\Big(n_{B}{\frac{A_{B}}{A_{R}}}\Big)+C_{3}\mathrm{log}_{10}C_{T}+C_{4}\mathrm{log}_{10}\Big({\frac{D}{\lvert\bm x\rvert}}\Big)-C_{5}\mathrm{log}_{10}D-C_{6}+10 \log_{10}\mathcal D(\bm x).
%\end{aligned}
\end{equation}

This model for wind turbine noise considers several contributing factors. The dominant contribution is the blade tip speed, defined as $V_\mathrm{tip}=\Omega R$, where $\Omega$ is the rotational speed and $R$ is the rotor radius (or blade length). Other relevant parameters include the rotor diameter $D$, the solidity of the wind turbine rotor $\sigma = n_B A_B/A_R$ and the thrust coefficient $C_T$. The model assumes spherical sound propagation, resulting in a decay rate of $1/\lvert\bm x\rvert$, where $\lvert\bm x\rvert$ is the distance to the source (i.e., the rotor center). These contributions are weighted by the coefficients proposed by \citet{hagg1990aerodynamic} via calibration with experiments, $C_1=63.3$, $C_2=11.5$, $C_3=2.5$, $C_4=20$, $C_5=10$ and $C_6=27.5$. Finally, we incorporate an enforced directivity function $\mathcal D(\bm x)=\cos\theta$ to recover the dipole radiation pattern aligned with the rotor axis.

The final wind farm noise emission is obtained by performing the logarithmic summation of the individual turbine contributions. To minimise the computational cost, no acoustic propagation methods are considered beyond spherical spreading. The final acoustic metric $\hat L_p$ used in the reward function \eqref{eq:reward_def} is defined as the maximum wind farm noise level within the control region:
\begin{equation}
    \hat L_p = \max_{\bm x\in\mathcal R} L_p^F(\bm x) = \max_{\bm x\in\mathcal R} 10\log_{10}\left(\sum_{i=1}^N 10^{L_{p,A}^i/10}\right). \label{eq:wind_farm_noise}
\end{equation}

To evaluate the maximum noise level within $\mathcal R$, the region is discretised using a regular $100\times100$ grid, and the maximum value over all grid points is selected. 

The proposed low-fidelity wind farm noise model can be coupled to \floris to estimate the acoustic impact with a computational cost similar to that of the flow simulator. However, \floris does not provide the rotor speed required to determine the blade tip speed in \eqref{eq:noise_model}, as the wind turbine is represented by a power curve. The rotational speed can instead be estimated by assuming that the wind turbine operates according to a standard variable-speed controller. Under this assumption, the rotor speed as a function of the local wind speed $\hat U$ is estimated to be $\Omega(\hat U)=\min\left(\lambda^*\hat U/R,\Omega_\mathrm{rated}\right)$,
where $\lambda^*=7.5$ denotes the optimal tip-speed ratio of the wind turbine. The proposed model is validated against the \ff aeroacoustic model in \autoref{app:noise_validation}, showing a modelling mismatch in the \SI{1}{\decibel} range.

\section{Training and analysis of the Noise Steering policy in \floris}
\label{sec:floris_results}
In this section, an RL agent is trained using \floris, along with the low-fidelity noise model developed in \autoref{sec:aeroacoustic}. The training process consisted of 300k steps (interactions with the solver \floris) divided in episodes of 500 steps. In each episode, a random steady wind speed is picked between 8 and 15~\ms, covering both under- and above rated wind conditions. The wind direction remains constant and aligned to the wind turbine rows throughout all episodes.

\begin{figure}[htbp]
    \centering
    \includegraphics[width=0.6\linewidth]{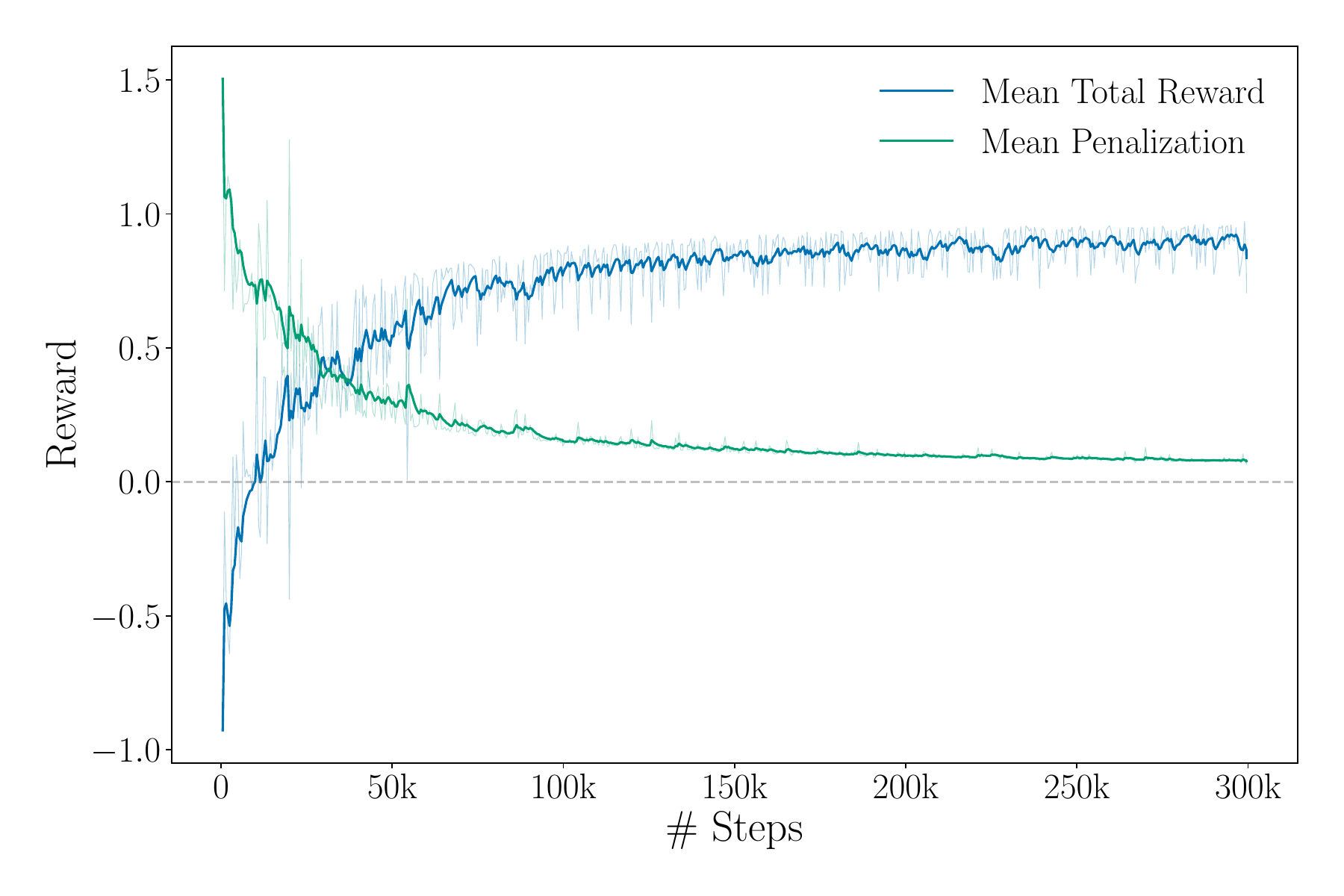}
    \caption{Average episode reward and penalisation terms during training. An episode corresponds to 500 steps. Bold curves are filtered over using an exponential smoothing of 0.8. }
    \label{fig:training_reward}
\end{figure}

\autoref{fig:training_reward} shows the average episode reward gathered by the agent throughout the training process. The penalisation components (i.e. noise threshold and action penalties) of the reward \eqref{eq:reward_def} are also displayed to illustrate the agent's ability to learn how to maximise power generation, while simultaneously reducing the acoustic impact of the wind farm. It should be noted that the penalisation terms do not reach zero because of the agent's exploration during training. In addition, the action-related penalisation terms cannot be reduced to zero throughout the episode.

To quantify the performance of the trained RL agent, we run one evaluation episode at each wind speed from 8 to 15 m/s in 0.5 m/s increments, and report the wind farm power output and noise emissions reached at the end of each episode. Three different control strategies are tested: 
\begin{itemize}
    \item \textbf{Noise Steering}. RL agent trained to maximise power generation subject to a noise constraint using the reward defined in \eqref{eq:reward_def}.  
    \item \textbf{Greedy control}. Conventional variable-speed wind turbine controller operating independently and in a decentralised manner. Each wind turbine aims to maximise its individual power generation.
    
    \item \textbf{Noise reduced operation, NRO}. Industry standard mode for noise reduction. The variable-speed controller of each wind turbine is derated by reducing the rated rotor speed to mitigate noise emissions at the expense of power production \citep{van2025sound}. In practice, this results in a curtailed power curve. A uniform derating across the wind farm is considered. To achieve a noise level of \SI{40}{\decibel} within the control region, the minimum uniform derating is determined for each wind speed by iteratively decreasing the power setpoints until the noise constraint is met. 
\end{itemize}

\begin{figure}[htbp]
\centering 
     \begin{subfigure}[t]{0.55\textwidth}
        \centering
        \includegraphics[width=0.8\linewidth]{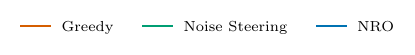}
    \end{subfigure}
    \centering
   \begin{subfigure}[t]{0.45\textwidth}
        \centering
        \includegraphics[width=\linewidth]{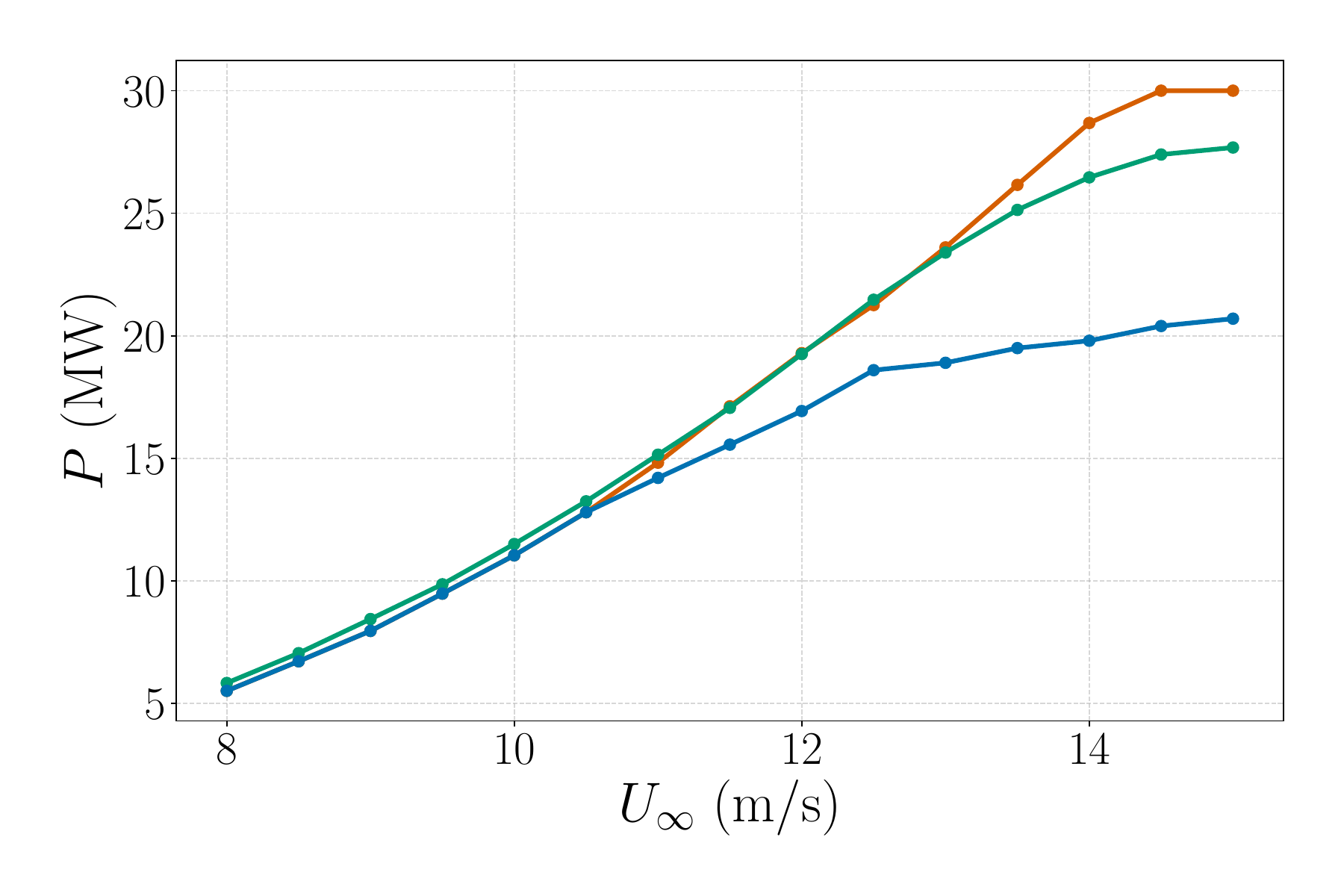} 
        \caption{Farm power.}
        \label{fig:T6_Floris_sweep_power}
    \end{subfigure}
    \begin{subfigure}[t]{0.45\textwidth}
        \centering
        \includegraphics[width=\linewidth]{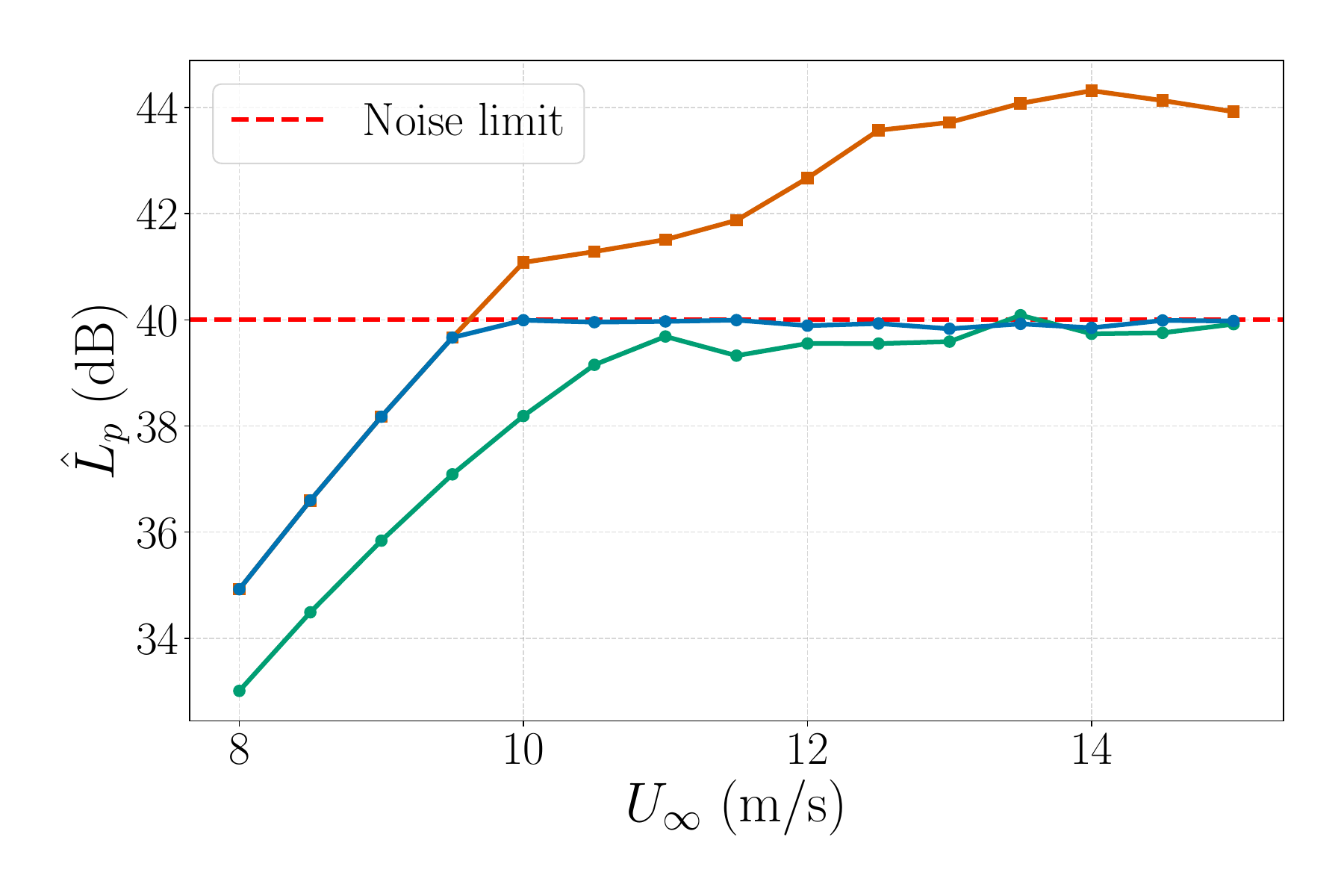} 
        \caption{Acoustic emissions at the control region.}
        \label{fig:T6_Floris_sweep_ospl}
    \end{subfigure}
    \caption{Power and noise emissions reached by the wind farm at the end of one episode for the complete wind speed range. Comparison of different control strategies.}
    \label{fig:T6_Floris_Sweep}
\end{figure}

\autoref{fig:T6_Floris_Sweep} illustrates the wind speed sweep of the wind farm power output and the noise level for the three control strategies. The \textit{greedy control} achieves the highest wind farm power output; however, it exceeds the noise constraint for wind speeds above 10~\ms. 
In contrast, the NRO control strategy satisfies \SI{40}{\decibel} threshold by design, but substantially limits wind farm power generation.
The uniform power setpoints commanded by the NRO strategy to the turbines range from \SI{4}{\mega\watt} at 10~\ms to \SI{3.4}{\mega\watt} at 15~\ms.
At 15~\ms, the NRO strategy produces approximately \SI{20}{\mega\watt}, corresponding to only 2/3 of the available power. This control strategy exhibits an average power deficit of 17\% relative to \textit{greedy control} over the wind speed sweep.

However, the trained \textit{Noise Steering} agent learns to exploit the interactions between wind turbines to significantly reduce the noise level within the control region, while maintaining it below the threshold $L_{p,\mathrm{thr}}$ across the complete wind speed range. At the same time, the agent maintains a wind farm power output close to that of the greedy policy, with limited power losses only above 13~\ms. Across the complete wind speed sweep, the \textit{Noise Steering} policy exhibits an average power deficit of 2\% relative to \textit{greedy control}. At the highest wind speed considered, the power deficit is 7.7\% while achieving a noise reduction of \SI{4}{\decibel}.

\begin{figure}[htbp]
  \centering  
     \begin{subfigure}[t]{0.55\textwidth}
        \centering
        \includegraphics[width=0.8\linewidth]{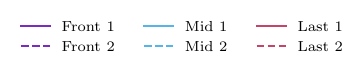}
    \end{subfigure}
    \centering
   \begin{subfigure}[t]{0.45\textwidth}
        \centering
        \includegraphics[width=\linewidth]{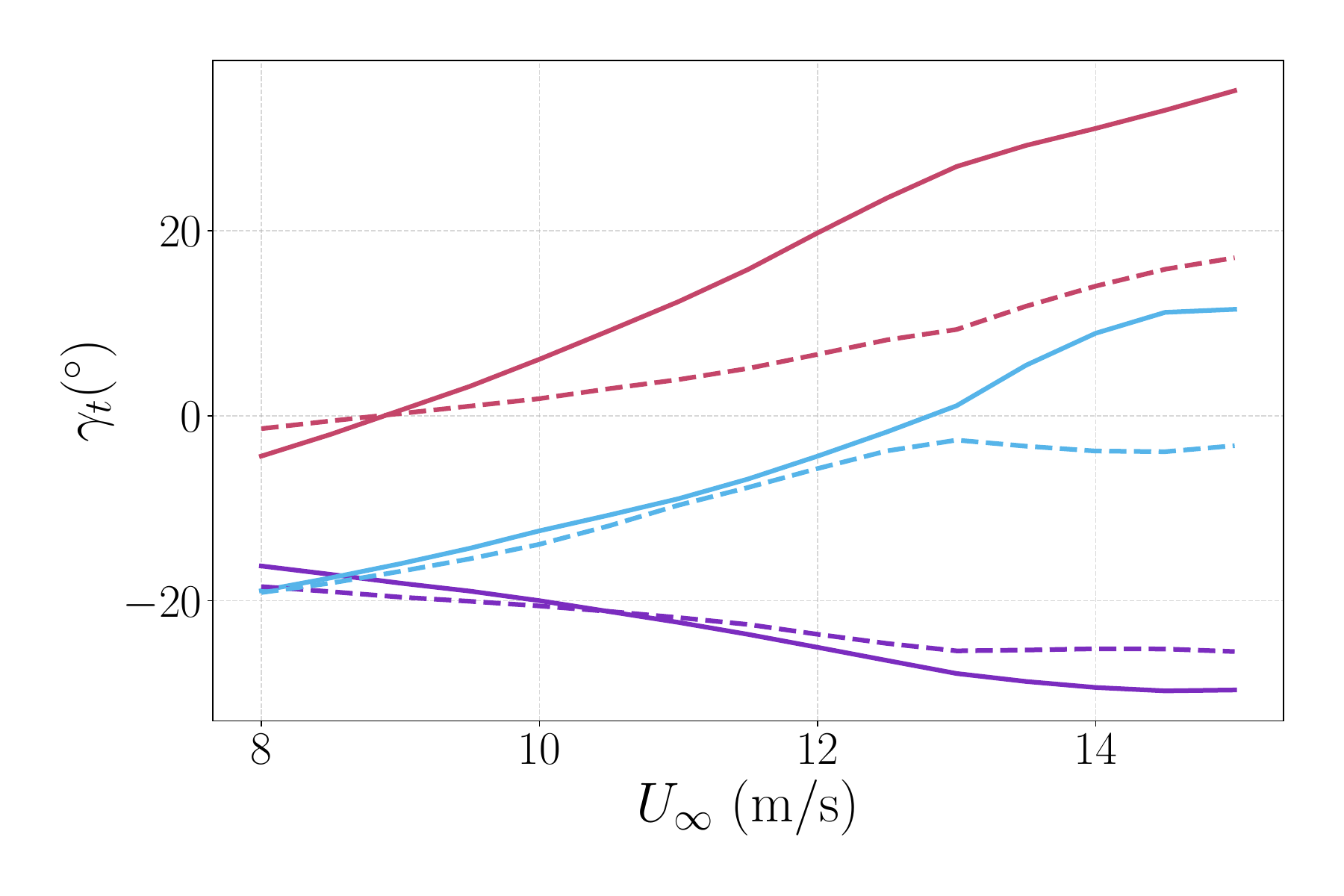} 
        \caption{Nacelle yaw.}
        \label{fig:Yaw_action}
    \end{subfigure}
    \begin{subfigure}[t]{0.45\textwidth}
        \centering
        \includegraphics[width=\linewidth]{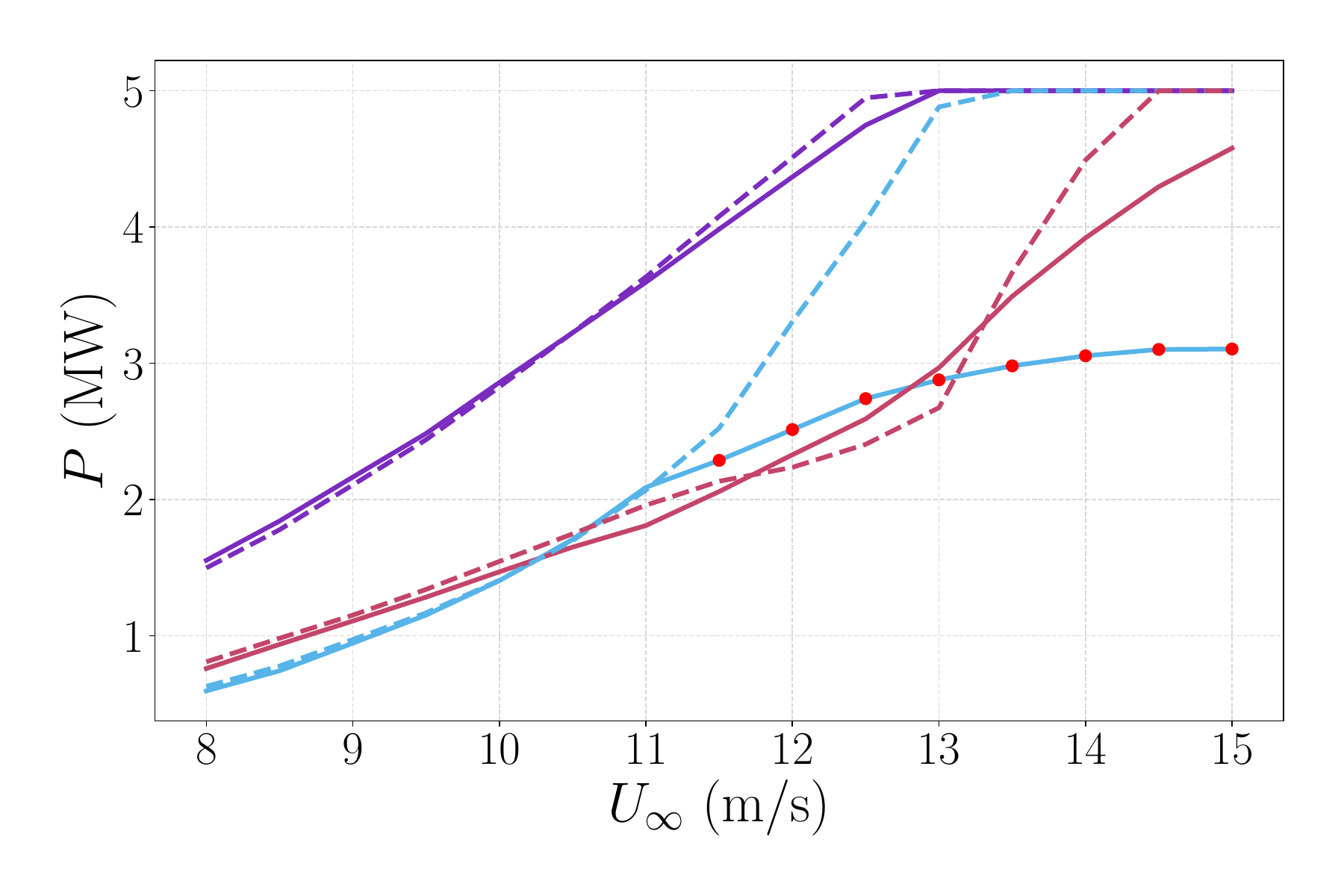} 
        \caption{Individual powers. \say{$\color{red}{\bullet}$} denotes when a power setpoint actively limits the power output of the turbine.}
        \label{fig:power_setpoint_action}
    \end{subfigure}
    \caption{Final commands of the \textit{Noise Steering} policy for the complete wind speed range. The wind turbine naming follows \autoref{fig:wf_layout} convention, where "1" stands for the row closer to the acoustic region and "2" for the other row of wind turbines.}
    \label{fig:T6_Floris_Actions}
\end{figure}

To investigate how the agent achieves this performance, we now look at the actions chosen for the same wind sweep at the end of each wind episode. \autoref{fig:Yaw_action} shows the yaw angles reached by the six wind turbines for the different wind speeds, while \autoref{fig:power_setpoint_action} presents the power generated by each turbine and indicates the turbines for which a power setpoint actively limits the power.

Firstly, for yaw control (\autoref{fig:Yaw_action}), at low wind speeds, the strategy is similar to  conventional power-maximizing wake steering~\citep{howlandWindFarmPower2019,monrocActorCriticAgents2023}. In fact, up to 11~\ms, the \textit{Noise Steering} agent achieves a slight increase in power output of approximately \SI{400}{\kilo\watt} relative to \textit{greedy control} (see \autoref{fig:T6_Floris_sweep_power}) as a result of this wake steering strategy. The front and mid turbines of both rows exhibit similar yaw angles around \SI{-20}{\degree}, while both last turbines face the wind (around \SI{0}{\degree} yaw). This strategy deviates from the wakes of the two front turbines to decrease the negative impact of their velocity deficit. 
At higher wind speeds, a similar power maximisation objective would breach the acoustic constraint. Therefore, the learned strategy consists of a low yaw angle (between \SI{-10}{\degree} and \SI{10}{\degree}) for both mid turbines and a positive yaw angle for both last two turbines (respectively, \SI{34}{\degree} and \SI{18}{\degree} at 15~\ms). This can be explained as the turbines preferentially yawing in the direction that aligns their rotor planes with the acoustic region, corresponding to a direction of reduced radiation due to the dipole-like directivity of the wind turbine noise.

Leveraging the dipole noise emission structure with the yaw angles is not sufficient in some cases and has to be combined with some curtailing using power setpoints (\autoref{fig:power_setpoint_action}). Lowering the power of a turbine by lowering its rotor speed will produce less noise, as exploited by the baseline NRO strategy. In this case, it is only necessary for the middle turbine of row 1, which is closest to the control region. It is required to reduce its power generation for wind speeds greater than 11.5~\ms. This substantial derating reduces both the power output and noise emission of the turbine, but also enables better wake recovery for the last turbine in row 1.

\begin{figure}[htbp!]
    \centering 
    \includegraphics[width=0.7\linewidth]{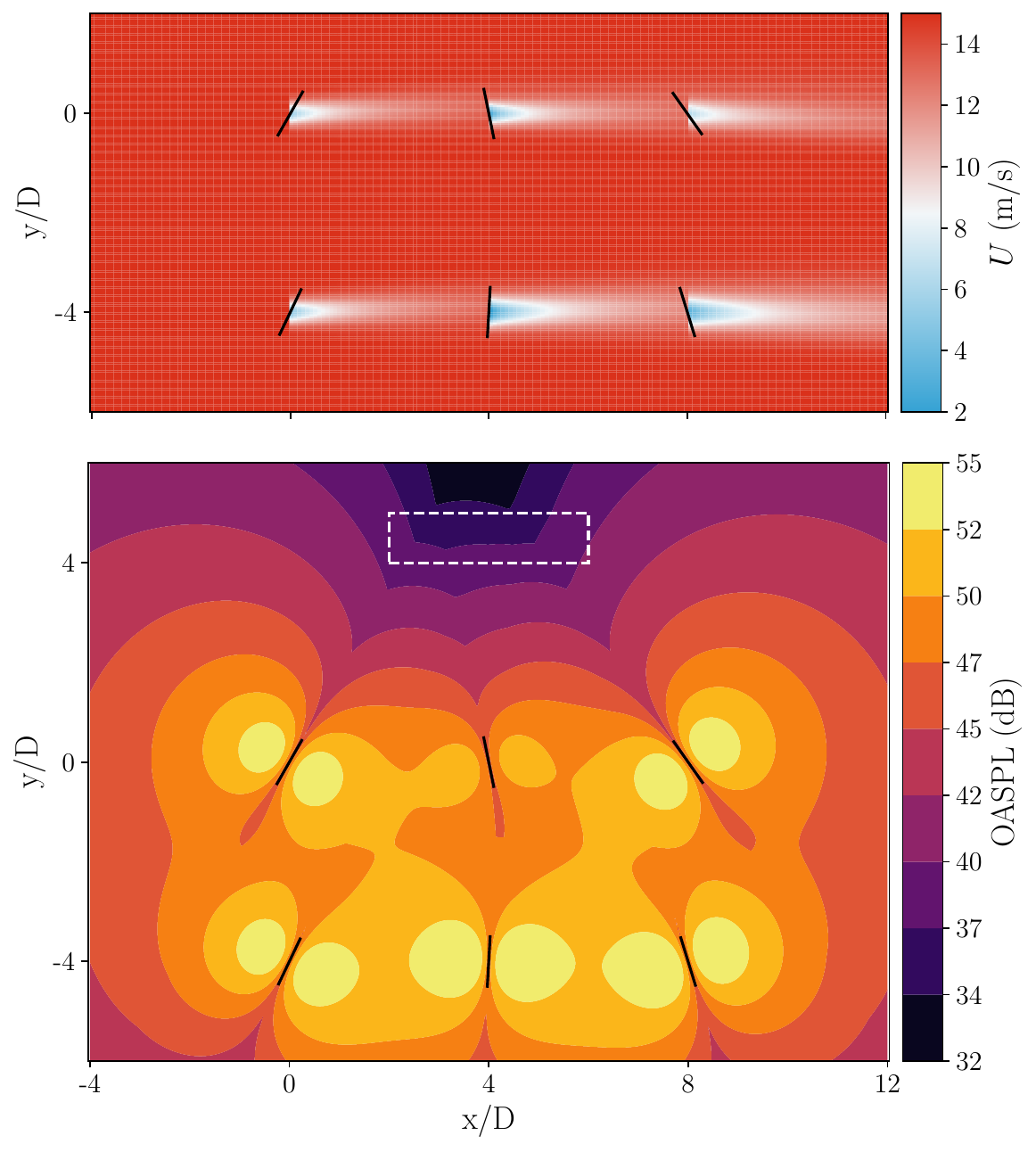}
    \caption{Wind farm final state reached by the RL policy for $U_\infty=15~\ms$ on \floris. Top: Wind speed field. Bottom: Acoustic field, control region denoted for reference.}
    \label{fig:floris_15ms}
\end{figure}

To further explain the \textit{Noise Steering} policy, the wind farm flow and acoustic fields are presented in \autoref{fig:floris_15ms} for the highest wind speed considered in the sweep, 15~\ms. 
\autoref{fig:floris_15ms} (bottom) shows that all turbines direct their rotor planes towards the acoustic region. This showcases how the directivity of the turbine noise emission can be exploited to control the noise in the desired area.
The noise emission is also visibly lower around the mid turbine of the first row compared to the mid turbine of the second row, due to the power curtailment described above. 
For upstream turbines, orienting its turbine plane towards the control area also produces a beneficial wake steering effect (\autoref{fig:floris_15ms} top), allowing higher power extraction for downstream turbines.

Overall, the \textit{Noise Steering} policy achieves power levels comparable to those of the \textit{greedy control} while substantially reducing noise emissions. 
When trying to control the noise level in a localised control area, leveraging the turbine noise directivity allows reducing noise emissions with limited loss of power production compared to the industry standard. These results indicate that collaborative wind farm control enables a more favourable trade-off between power production and noise reduction.

\section{Evaluation in FAST.Farm: robustness to model mismatch and unsteady inflow}
\label{sec:FF_results}
In this section, the \textit{Noise Steering} policy obtained in \autoref{sec:floris_results} is deployed in a \ff environment to assess its validity using a higher-fidelity wind turbine solver. \autoref{sec:ff_integration} explains how to integrate the policy trained in \floris so that it can be used directly in a \ff environment. Afterwards, the policy is evaluated under constant wind speed in \autoref{sec:ff_sweep} and a turbulent wind speed profile in \autoref{sec:ff_turb}.

\subsection{\ff integration}
\label{sec:ff_integration}
The RL policy acts with a $\Delta t_\text{RL}=\SI{3}{\second}$, however, the \ff variable-speed controller actuates continuously during the simulation. To couple these two time scales, we consider the RL action to set a fixed command or condition for the variable-speed controller. Therefore, $\Delta\bm{\gamma_t}$ sets a new yaw command for the yaw controller and $\Delta\mathbf{P_S}$ sets a new rated power that defines the variable-speed controller. 

To ensure physically feasible yaw actuation, a yaw-rate limiter is applied to the RL yaw command. This limiter constrains the yaw rate to a maximum of \SI{0.3}{\degree\per\second}, consistent with industry standards.

The variable-speed controller implemented in the WFCRL \ff interface is also modified to allow for turbine derating. Following the Region II $K\Omega^2$ control law \citep{mulders2018delft}, the rated rotor speed $\Omega_\mathrm{rated}$, the rated generator torque $T_{g,\mathrm{rated}}$, and the rated power are modified according to the imposed power setpoint. The resulting rated rotor speed and generator torque are given by
\begin{equation}
\label{eq:rpm_derated}
\Omega_\mathrm{rated}^{'}=
\left(\frac{P_S}{P_\mathrm{rated}}\right)^\frac{1}{3}
\Omega_\mathrm{rated},
\end{equation}
and
\begin{equation}
\label{eq:torque_derated}
T_{g,\mathrm{rated}}^{'}=
\left(\frac{P_S}{P_\mathrm{rated}}\right)^\frac{2}{3}
T_{g,\mathrm{rated}}.
\end{equation}

Thus, the RL power setpoint command is translated into a modified set of rated operating conditions for the variable-speed controller. This modification results in different generator torque and blade-pitch actuation, thereby implementing turbine power derating. 

Regarding wind turbine noise prediction, the integrated aeroacoustic module in \ff is employed. The control region is discretised using an $8\times2$ grid, and the acoustic signal is processed using exponential smoothing to eliminate the wind turbine noise amplitude modulation. Increasing the number of observers greatly increases the computational cost of the simulation, hence the choice for less observers than in \floris.

\subsection{Steady wind speed comparison}
\label{sec:ff_sweep}
In this section, the \textit{Noise Steering} policy is evaluated under steady wind speed in the \ff environment. This assessment examines the ability of a policy trained in \floris to perform in a different wind turbine solver while accounting for the transient wake dynamics resolved by \ff.

\begin{figure}[htbp]
    \centering 
     \begin{subfigure}[t]{0.75\textwidth}
        \centering
        \includegraphics[width=\linewidth]{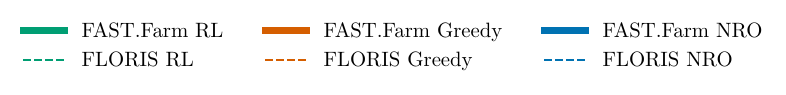}
    \end{subfigure}
    \centering
   \begin{subfigure}[t]{0.45\textwidth}
        \centering
        \includegraphics[width=\linewidth]{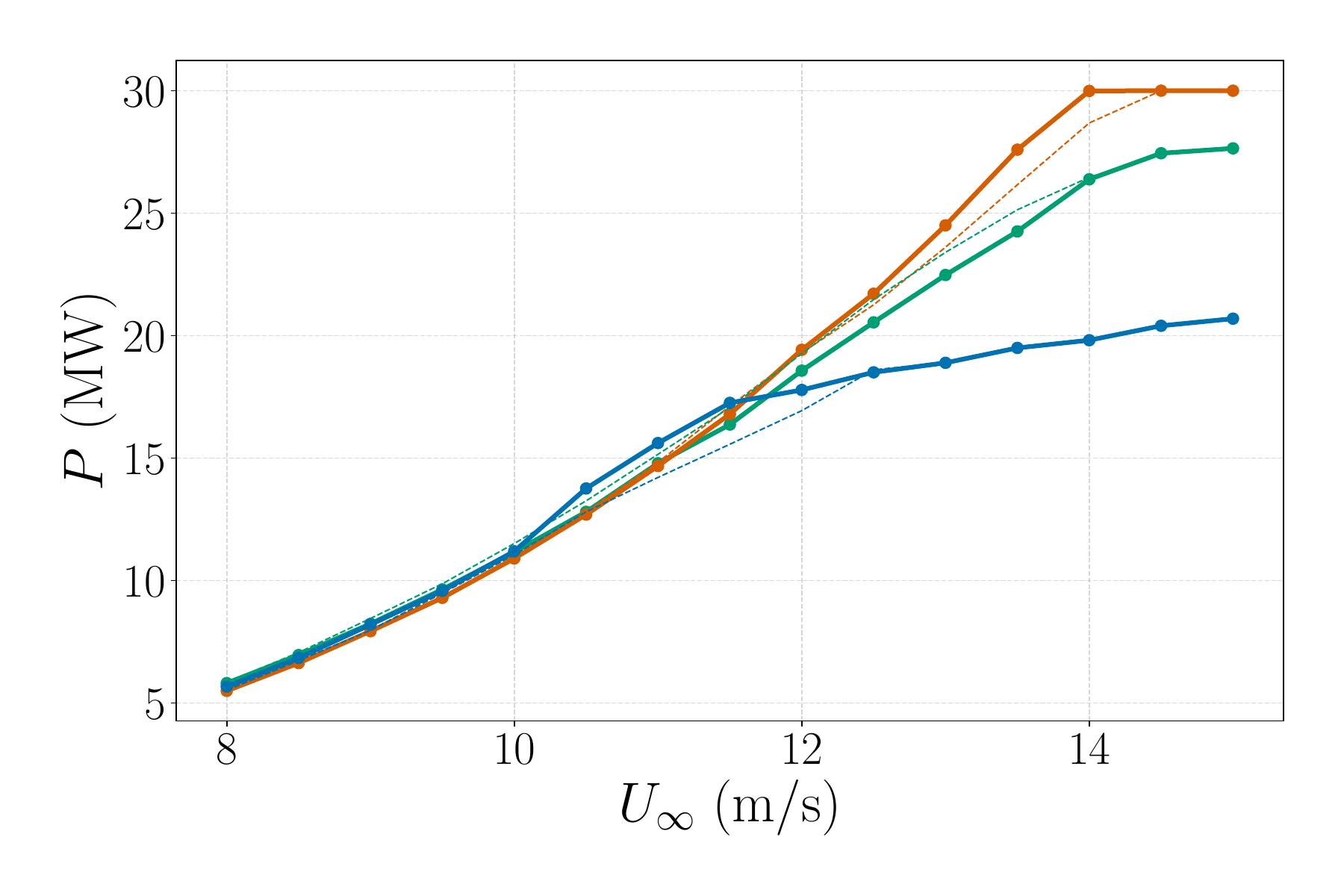} 
        \caption{Farm power.}
        \label{fig:FF_sweep_power}
    \end{subfigure}
    \begin{subfigure}[t]{0.45\textwidth}
        \centering
        \includegraphics[width=\linewidth]{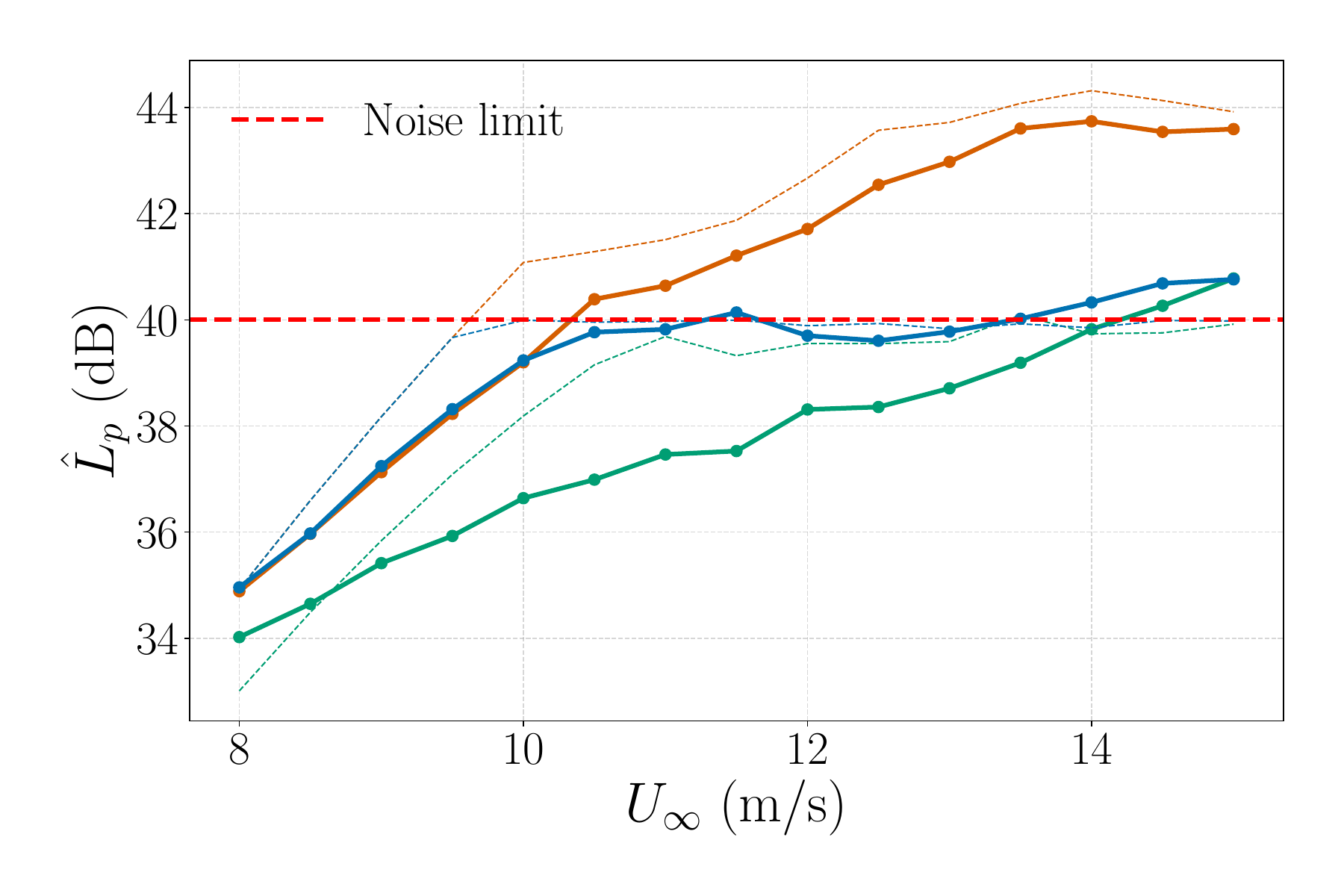} 
        \caption{Acoustic emissions at the control region.}
        \label{fig:FF_ospl}
    \end{subfigure}
    \caption{Comparison of the three different control strategies on \ff and \floris for the complete wind speed range.}
    \label{fig:FF_sweep}
\end{figure}

\autoref{fig:FF_sweep} illustrates the comparison of the three studied control strategies under the complete wind speed range for both wind turbine solvers. The NRO control strategy and the \textit{Noise Steering} RL agent are evaluated directly in \ff without further adjustment.

The power generation for the \textit{greedy control} and NRO strategies are closely matched between the two simulators (see \autoref{fig:FF_sweep_power}), thanks to the calibration described in \autoref{sec:calibration}. Both strategies produce noise levels close to their \floris results within the \SI{1}{\decibel} modelling error identified in \autoref{app:noise_validation} for the aeroacoustic model. NRO exceeds the noise threshold above 14\ms, due to the uniform derating levels being optimised on \floris and transferred as is.

The \textit{Noise Steering} strategy achieves a similar power output when deployed in \ff as in \floris (see \autoref{fig:FF_sweep_power}). Above 12\ms, the power deficit compared to greedy is greater but still outperforms the NRO baseline. 
In terms of noise emissions, \autoref{fig:FF_ospl} shows that the RL policy satisfies the noise constraint at all wind speeds except the highest considered. However, the \ff evaluation results in lower noise levels than those predicted by \floris, with a difference of up to \SI{2.5}{\decibel} at 11~\ms. 
This indicates that the policy might be sub-optimal in \ff due to model mismatch. However, contrary to model-based methods, RL methods are able to continue learning across simulators \citep{taylor2009transfer,Ebrahimi_2026,monrocFineTuningWake2024}. Therefore, policies resulting in higher power while being closer to the noise limit could be found through fine-tuning in a future study.

Note that the RL controller is a closed-loop controller.  Hence, when the \textit{Noise Steering} policy trained in \floris is deployed in the \ff environment, the differences arising from the modelling gap modify the state-action transitions. Consequently, the policy may produce different final values of $\bm{\gamma}_t$ and $\mathbf{P}_S$. This is further discussed in \autoref{app:ff_floris}, validating that the policy results in similar commands on both simulators.

\subsection{Turbulent wind assessment}
\label{sec:ff_turb}

Finally, the \textit{Noise Steering} policy is evaluated under an unsteady wind inflow to determine its ability to adapt to turbulent and fluctuating wind conditions. 
% Although the policy is trained under constant wind conditions, it is evaluated under turbulent inflow, as is common practice in reinforcement-learning studies of wind-turbine control; previous works have reported similar control performance under both conditions~\citep{SOLER2024123502,frutos2025enhancing,wfcrl}.

The turbulent wind speed profile is generated by combining an unsteady wind speed profile $\mathcal U(t)$ with a shear layer characterized by an exponent $\alpha_\text{PL}=0.2$ and turbulent fluctuations generated using the stochastic turbulence inflow simulator TurbSim~\citep{jonkman2006turbsim}. The inflow wind speed is defined as follows,  
\begin{equation}
\label{eq}
u(y,z,t) = \mathcal U(t)\left(\frac{z}{z_\mathrm{ref}}\right)^{\alpha_\text{PL}} + \sigma(\mathcal U(t))u'(y,z,t), 
\end{equation}
where $u'(y,z,t)$ is the normalized TurbSim inflow and a constant turbulence intensity (TI) approach is used to scale the turbulent fluctuations, with $\sigma=\mathcal U(t)~\text{TI}$. A turbulence intensity of $\text{TI}=\SI{6}{\percent}$ is considered. \ff convects the resulting inflow field at a constant velocity $u_\text{ref}=\text{mean}(\mathcal U(t))$.

\begin{figure}[htbp]
    \centering
    \includegraphics[width=0.75\linewidth]{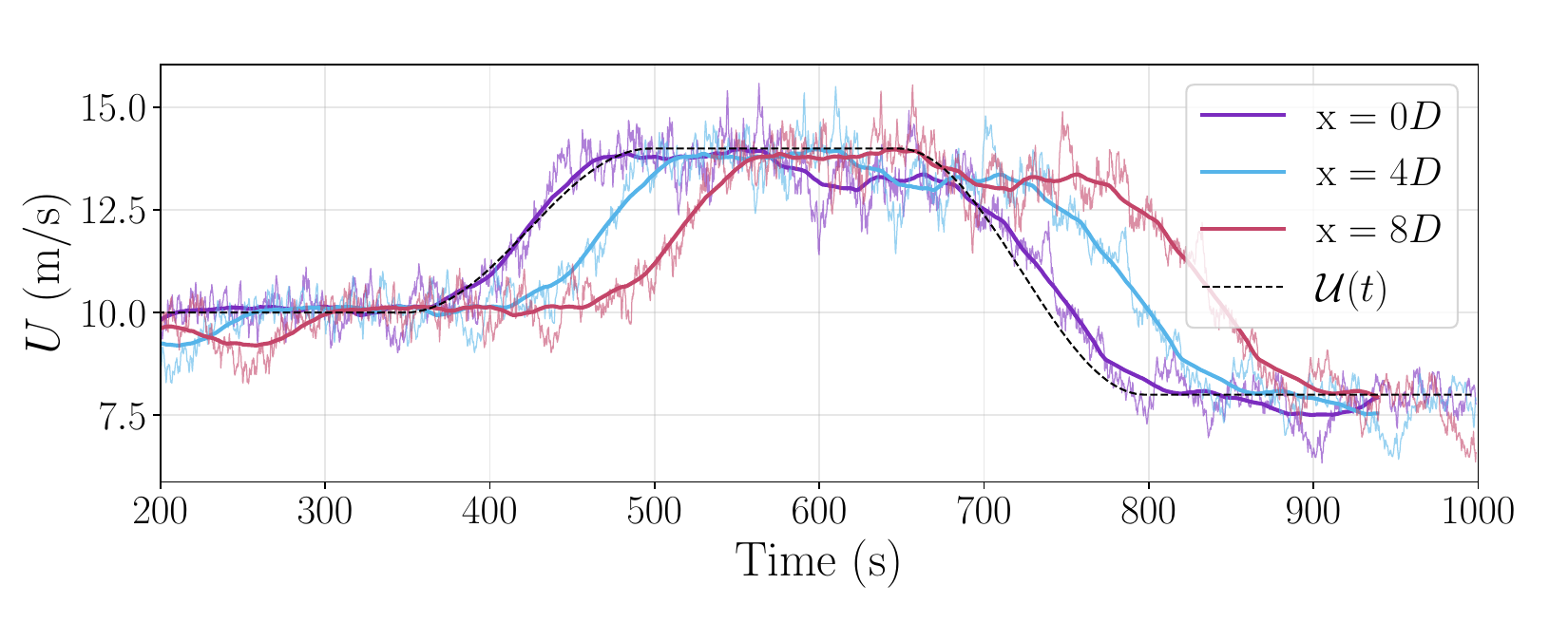}
    \caption{Turbulent and unsteady wind speed profile. Ambient wind displayed at hub height for the three wind turbine positions of the first row (front, mid and last). Bold curves are filtered using a \SI{60}{\second} moving average.}
    \label{fig:turbulent wind}
\end{figure}

\autoref{fig:turbulent wind} illustrates the temporal evolution of the inflow wind speed at hub height and at the first row of turbines. The evolution of the wind speed is also shown in the different positions $x/D$ within the wind farm, together with the original profile $\mathcal U(t)$. The prescribed unsteady profile starts at 10~\ms, increases to 14~\ms, and finally decreases to 8~\ms. This profile therefore evaluates the ability of the policy to adapt to varying wind speeds across different operating conditions and to the effects of turbulence.

\begin{figure}[htbp]
    \centering 
     \begin{subfigure}[t]{0.55\textwidth}
        \centering
        \includegraphics[width=\linewidth]{figures/Floris_sweep_legend.pdf}
    \end{subfigure}
    \centering
   \begin{subfigure}[t]{0.45\textwidth}
        \centering
        \includegraphics[width=\linewidth]{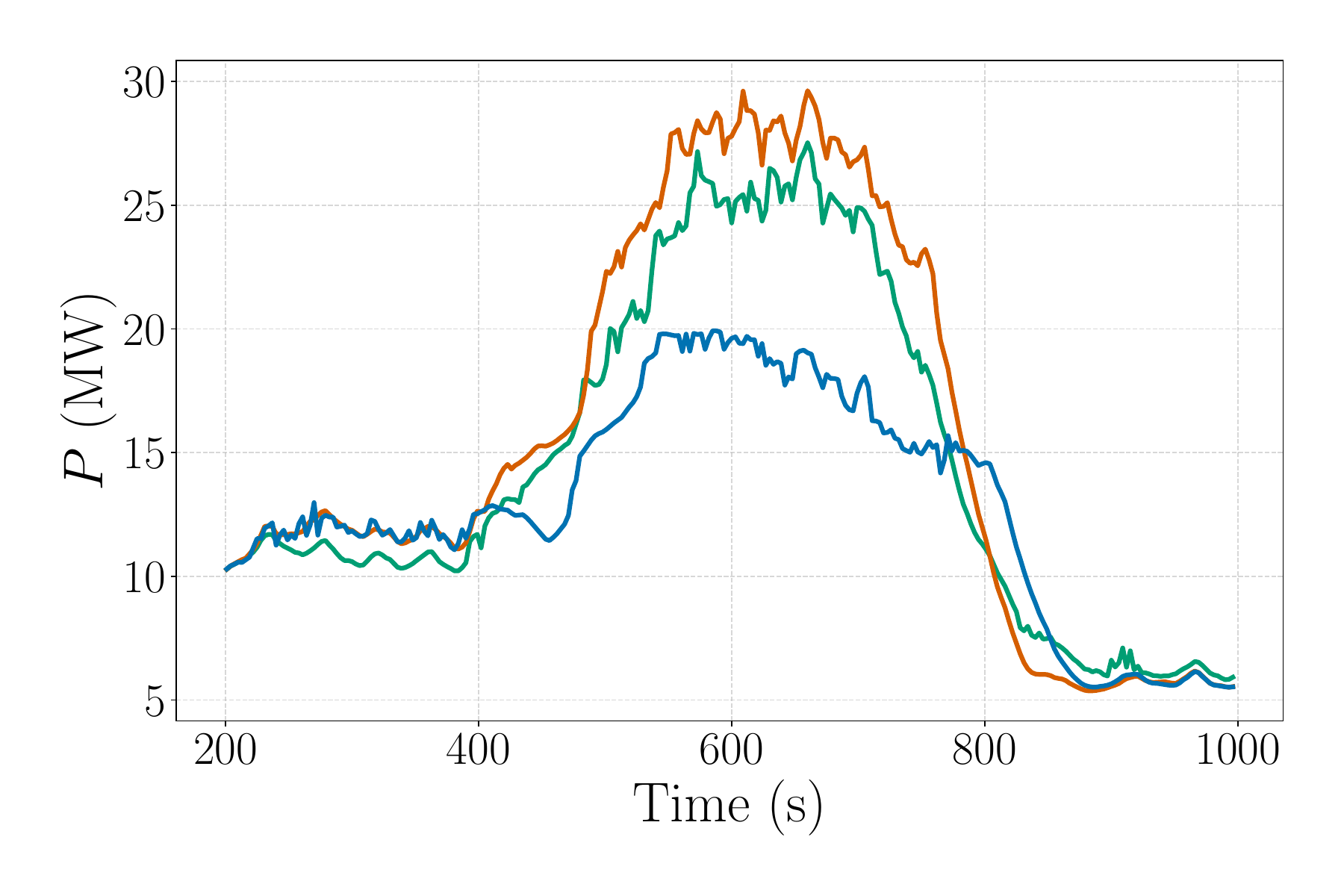} 
        \caption{Farm power.}
        \label{fig:FF_turb_power}
    \end{subfigure}
    \begin{subfigure}[t]{0.45\textwidth}
        \centering
        \includegraphics[width=\linewidth]{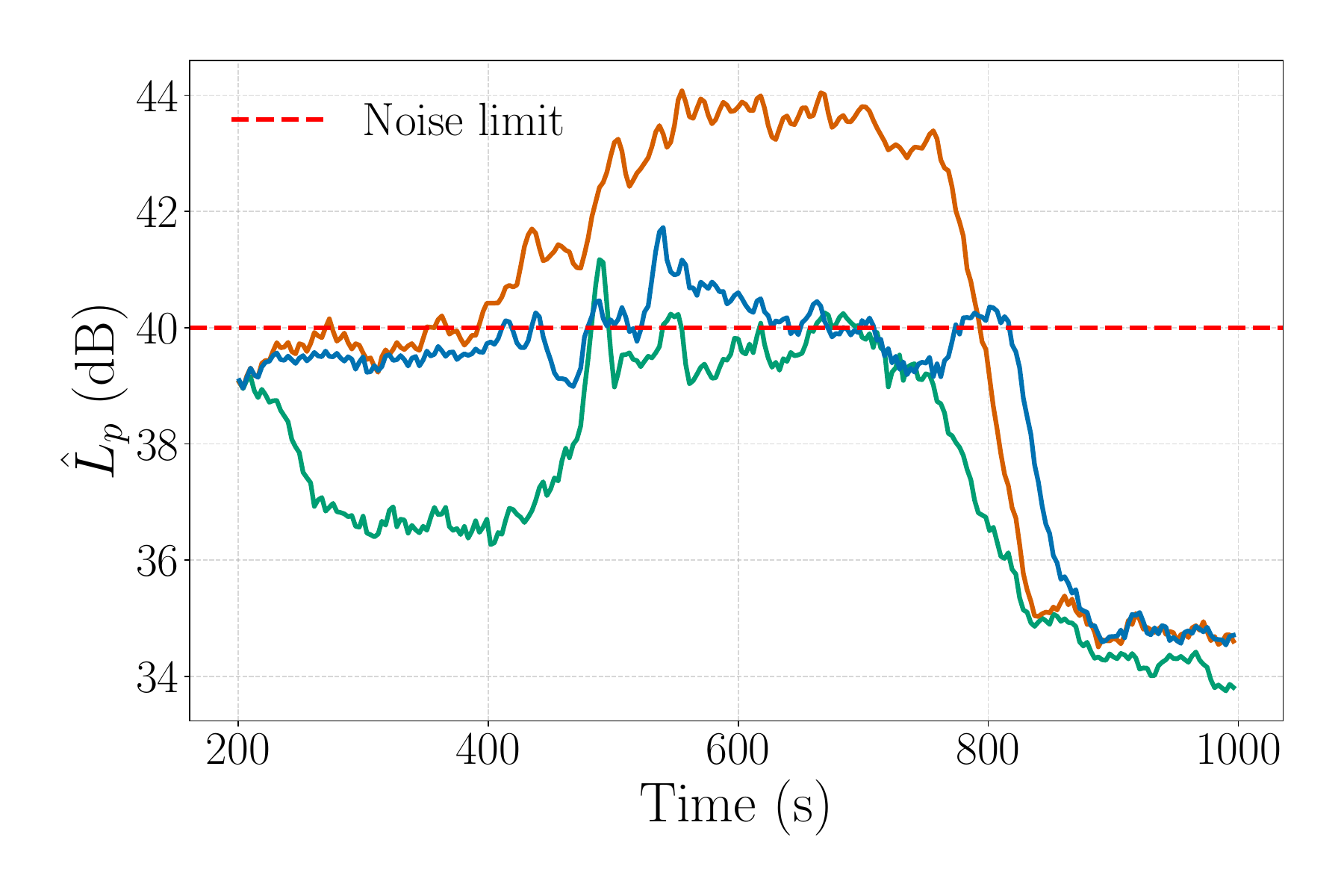} 
        \caption{Acoustic emissions at the control region.}
        \label{fig:FF_turb_ospl}
    \end{subfigure}
    \caption{Comparison of the three different control strategies on \ff for an unsteady inflow.}
    \label{fig:FF_turb}
\end{figure}

The three control strategies studied are evaluated under this turbulent wind inflow. Before activating control policies, an initialization period of \SI{200}{\second} is included to allow the wakes to develop fully. \autoref{fig:FF_turb} illustrates the temporal evolution of wind farm power generation and noise emissions for the three control strategies.

In general, the three control strategies behave similarly as observed under steady inflow
in \autoref{fig:FF_sweep}, showing that the \textit{Noise Steering} policy is capable of adapting to the unsteady inflow. 
During complete evaluation, \textit{greedy control} exceeds the noise limit during \SI{53.0}{\percent} of the time, with peaks of up to \SI{44.1}{\decibel}, and NRO during \SI{30.5}{\percent} of the time. In contrast, the \textit{Noise Steering} policy remains below the limit during \SI{91.7}{\percent} of the evaluation, while producing only \SI{7.4}{\percent} less power than \textit{greedy control}, compared to \SI{18.1}{\percent} for NRO.

The differences between the strategies are greatest on the 14\ms plateau (around 500 to 700\si{\second}). \textit{Greedy control} produces about \SI{28}{\mega\watt} at the cost of exceeding the noise limit by up to \SI{4}{\decibel}. In comparison, NRO reduces the power to about \SI{19}{\mega\watt} and still momentarily
exceeds the limit by almost up to \SI{2}{\decibel}. The \textit{Noise Steering} policy produces about \SI{25}{\mega\watt}, i.e.
a power deficit of approximately \SI{10}{\percent} relative to \textit{greedy control}, while satisfying the noise constraint. 
% These values are close to those obtained under steady inflow at 14\ms (\autoref{fig:FF_sweep}). 
The noise limit is only shortly exceeded by the \textit{Noise Steering} policy at the beginning of the 14\ms-wind speed plateau, reaching \SI{41.0}{\decibel}. The policy then reacts quickly and reduces the sound level to \SI{39.4}{\decibel} in 2 RL steps, i.e., 6 seconds.

This turbulent inflow further increases model-mismatch compared to the training simulator. Wake recovery is enhanced by ambient turbulence, which reduces the added benefit of wake steering~\citep{doekemeijerFieldExperimentOpenloop2021}. 
For example, at the 10\ms plateau, the \textit{Noise Steering} policy appears sub-optimal as it
produces less power than \textit{greedy control} and NRO, whereas it slightly outperformed both under steady inflow.

An animated visualization of the velocity field in the wind farm, and the noise level in the control area, is provided in Supplementary Material. This allows to visualize the Noise Steering control under this unsteady wind field.

\section{Conclusion}
\label{sec:conclusions}
This work introduced \textit{Noise Steering}, a farm-level control strategy that jointly exploits yaw misalignment and power derating to maximise wind farm power under a localised noise constraint. In this work, we chose Reinforcement Learning as a method to perform \textit{Noise Steering}, as it has the potential to provide a model-free closed loop controller.
% However, we also mention that \textit{Noise Steering} could be performed using other conventional control methods, such as Model Predictive Control. 

A low-fidelity aeroacoustic model, combining the analytical model of \citet{hagg1990aerodynamic} with an enforced dipole directivity, is coupled to the steady-state \floris simulator to enable RL training. 
Farm noise emissions are thereby estimated at the same order of computational cost as the flow field, making RL training tractable. 
A PPO agent is trained on a six-turbine farm to maximise power while keeping the sound level below \SI{40}{\decibel} in a control region located $4D$ from the closest turbine row.

The learnt policy satisfies the noise constraint throughout the 8-15~\ms range, with an average power deficit of only 2\% relative to \textit{greedy control} and a maximum of 7.7\% at the highest wind speed, compared to an average 17\% for the baseline NRO. 
The Noise Steering agent combines two complementary mechanisms. 
At low wind speeds, it reproduces conventional power-maximising wake steering. 
At higher wind speeds, it orients the rotor planes towards the control region, so that the dipole-like directivity of the turbine noise radiates away from it. 

Deployed without retraining in the higher-fidelity, time-resolved \ff solver under steady inflow, the policy exhibits similar power performance and remains below the noise limit at all but the highest wind speed. This indicates that the calibrated steady-state training environment is representative enough for transfer to the dynamic environment.
However, the transferred policy appears too conservative and additional power could be produced under the noise constraint. 
This is a consequence of model-mismatch between the simulators, which could hence be mitigated in future studies by enabling the RL agent to continue learning online in the higher-fidelity environment.

The policy is further evaluated under an unsteady and turbulent wind inflow. Although trained exclusively under steady inflow, the Noise Steering policy manages to respect the noise constraint under varying wind conditions. It exhibits only transient noise constraint violations while retaining 92\% of greedy power, thus improving conventional NRO power generation by 10\%.
% Violations of the noise constraint are limited to short transients during the wind speed ramps, of at most 1 dB, which is within the modelling error of the aeroacoustic model. 
% Under turbulent inflow, the power gains from wake steering at low noise-constrained wind speeds are reduced, while the behaviour at high wind speeds remains close to that found under constant inflow.

These results are obtained for a single wind direction, a fixed control region, and an acoustic model that neglects propagation effects such as ground reflection, refraction, and atmospheric absorption. Including unsteady inflow during training could further improve performance. The low-fidelity environment is also pre-calibrated to reduce model-mismatch with the higher-fidelity environment. The impact of relaxing those hypotheses should be assessed in future work.

 Given the energy production currently lost to acoustic curtailment \citep{Willis2023Curtailment}, we believe that wind farm control that takes advantage of wake interactions and acoustic directivity is a highly promising direction.

\section{Acknowledgement}
This research has received funding from the European Union (ERC, Off-coustics, project number 101086075). Views and opinions expressed are, however, those of the authors only and do not necessarily reflect those of the European Union or the European Research Council. Neither the European Union nor the granting authority can be held responsible for them.

Martín de Frutos and Esteban Ferrer acknowledge the funding received by the Comunidad de Madrid according to Orden 5021/2025, issued by the Consejero de Educacion, Ciencia y Universidades, which announces grants for the hiring of predoctoral research personnel in training for the year 2024.

Martín de Frutos gratefully acknowledges the Universidad Politécnica de Madrid (www.upm.es) for providing computing resources on Magerit Supercomputer.

This work is partially carried out in the framework of AI-NRGY (Grant ref.: ANR-22-PETA-0004).

\printcredits
\bibliographystyle{elsarticle-num-names}
\bibliography{references}

\newpage
\appendix
\section{RL hyperparameters}
\label{app:hyperparameters}

In this appendix, we detail the parameters chosen for the reward function and the PPO algorithm, see \autoref{tab:rl_hyperparameters}. 
These are chosen by trial and error. In the reward function, the action penalisation terms are kept as small as possible while limiting oscillations in the control policy. Higher penalties can result in excessively slow actions during operation.

\begin{table}[htbp]
    \centering
    \caption{Reward and PPO hyperparameters used for RL training.}
    \label{tab:rl_hyperparameters}
    \begin{tabular}{lll}
        \toprule
         & \textbf{Parameter} & \textbf{Value} \\
        \midrule
        \multirow{4}{*}{Reward}
        & $\alpha$ & 0.15 \\
        & $\beta_1$ & 0.002 \\
        & $\beta_2$ & 0.1 \\
        & $\beta_3$ & 0.0025 \\
        \midrule
        \multirow{9}{*}{PPO}
        & Learning rate & 0.0003 \\
        & Critic learning rate & 0.0006 \\ 
        & Discount factor $\gamma$ & 0.7 \\
        & GAE parameter $\lambda$ & 0.95\\
        & Entropy coefficient & 0.01 \\
        & Value function coefficient & 0.5 \\
        & Number of epochs & 4 \\
        & Minibatch size & 32 \\
        & Rollout length & 128 \\
        \midrule
        \multirow{1}{*}{RL agent}
 & \makecell[l]{Neural network size \\ (actor and critic)} & [64,64] \\
        \midrule
        \multirow{1}{*}{Environment}
        &Episode length & 500 \\
        &Wind speed range & [8,15]~\ms\\
        \bottomrule
    \end{tabular}
\end{table}
 
\section{Noise model validation}
\label{app:noise_validation}

The low-fidelity wind turbine noise model developed in \autoref{sec:aeroacoustic} is validated against the semi-empirical method implemented in \ff under three different operating conditions. To assess the complete modelling chain, three modelling approaches are compared. First, the mid-fidelity \ff + BPM approach is considered as the ground truth. Second, the effect of changing only the aeroacoustic model is evaluated using the wind farm operating conditions provided by \ff, resulting in the \ff + \citet{hagg1990aerodynamic} approach. Finally, the lower-fidelity approach used for training, namely \floris + \citet{hagg1990aerodynamic}, is evaluated.

The noise model is evaluated for three different operation conditions, ranging various wind speeds, yaw configurations and power setpoints. The specific conditions are described in \autoref{tab:noise_validation_cases}. \autoref{fig:Noise_validation} presents a comparison of the three modelling approaches for each of the considered operating conditions. For the sake of simplicity, the validation is assessed for the wind farm illustrated in \autoref{fig:wf_layout} considering only the first row of wind turbines. The results show that the low-fidelity approach captures both the directivity of the wind turbine noise and the corresponding amplitude levels under the different operating conditions.

\begin{table}[htbp]
    \centering
    \caption{Wind farm operating conditions for the noise validation cases shown in \autoref{fig:Noise_validation}. The difference between the higher fidelity model (\ff + BPM) and the lower fidelity model (\floris+Hagg) noise levels in the control region, $\Delta\hat L_p$,  is also reported.}
    \begin{tabular}{c | c c c || c}
        \toprule
       % \textbf{Case} & $U_\infty$ (m/s)& $\bm{\gamma_t}$ ($^\circ$) & $\mathbf {P_S}$ (MW) & $\Delta\hat L_p$ (dB)\\
       \textbf{Case} & $U_\infty\,(\ms)$ & $\bm{\gamma_t}\,(\si{\degree})$ & $\mathbf{P_S}\,(\si{\mega\watt})$ & $\Delta\hat{L}_p\,(\si{\decibel})$\\
        \midrule
        1 & 10.0   & $[30, 30, 0]$   & $[5, 5, 5]$      & 0.74 \\
        2 & 11.4   & $[0, 0, 0]$     & $[5, 5, 5]$      & -0.61 \\
        3 & 10.0   & $[0, -10, 20]$  & $[3, 3, 5]$      & 0.57 \\
        \bottomrule
    \end{tabular}
    \label{tab:noise_validation_cases}
\end{table}

\autoref{tab:noise_validation_cases} also reports $\Delta\hat L_p$, defined as the difference in $\hat L_p$, corresponding to the maximum noise level within the control region, between the higher- and lower-fidelity modelling approaches. The \floris + \citet{hagg1990aerodynamic} approach remains within ~\SI{1}{\decibel} of the \ff implementation for $\hat L_p$ across all evaluated operating conditions.

\newsavebox{\noisetablebox}
\sbox{\noisetablebox}{%
  \setlength{\tabcolsep}{2pt}
  \renewcommand{\arraystretch}{1}
  \begin{tabular}{cccc}
    & \textbf{Case 1} & \textbf{Case 2} & \textbf{Case 3} \\
    \rotatebox{90}{\footnotesize\shortstack{\textbf{FAST.Farm}\\\textbf{\& BPM}}} &
    \includegraphics[width=0.27\textwidth]{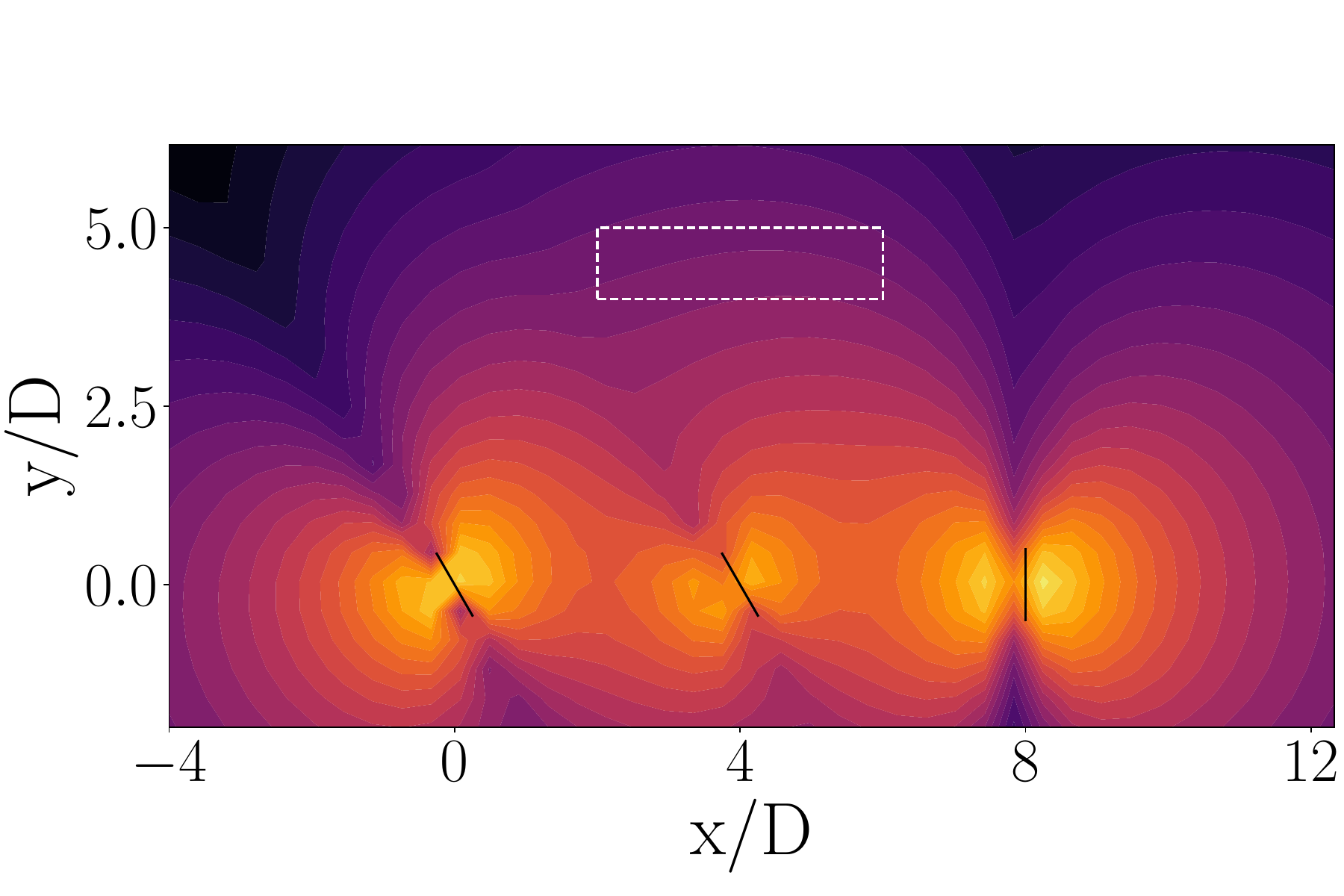} &
    \includegraphics[width=0.27\textwidth]{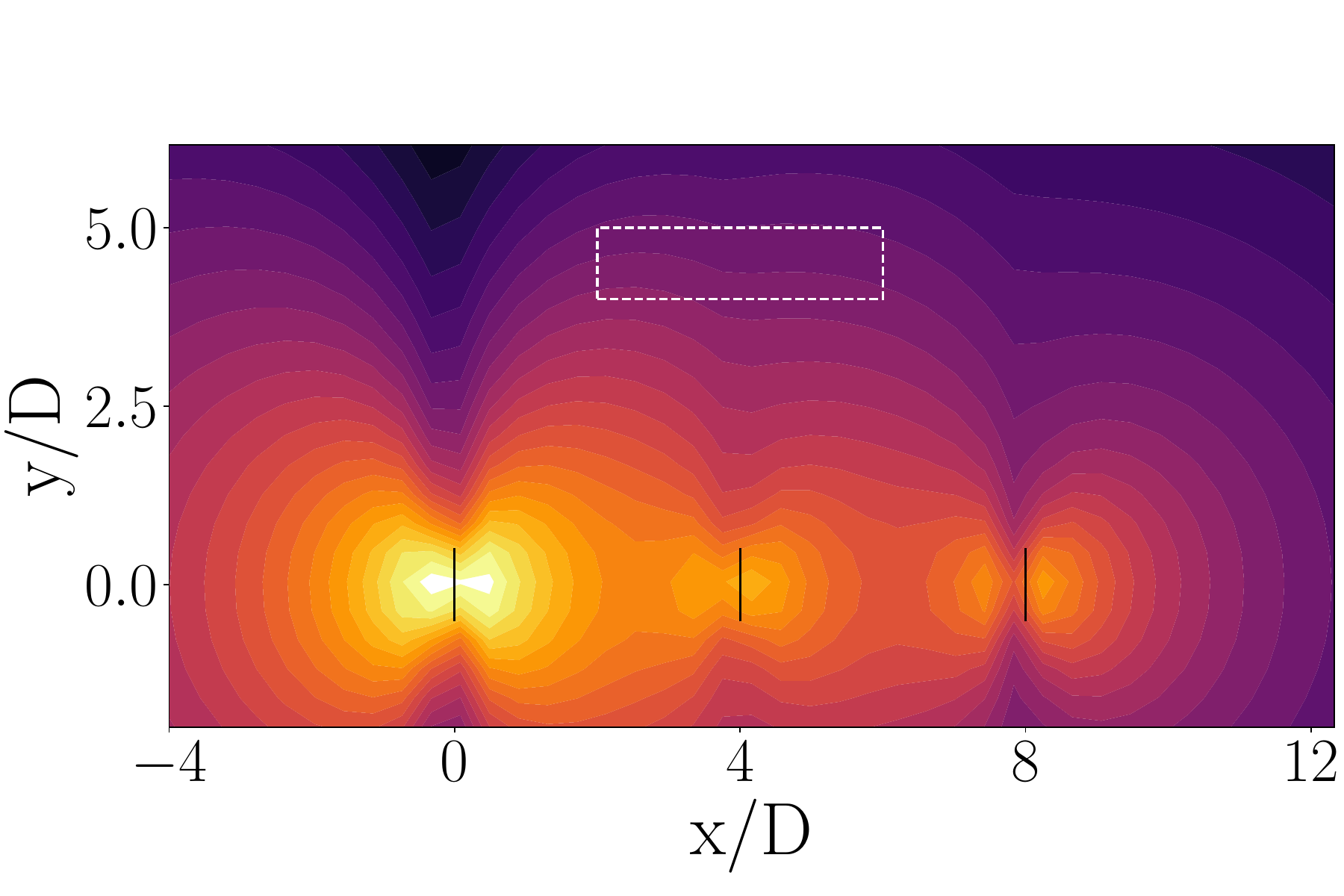} &
    \includegraphics[width=0.27\textwidth]{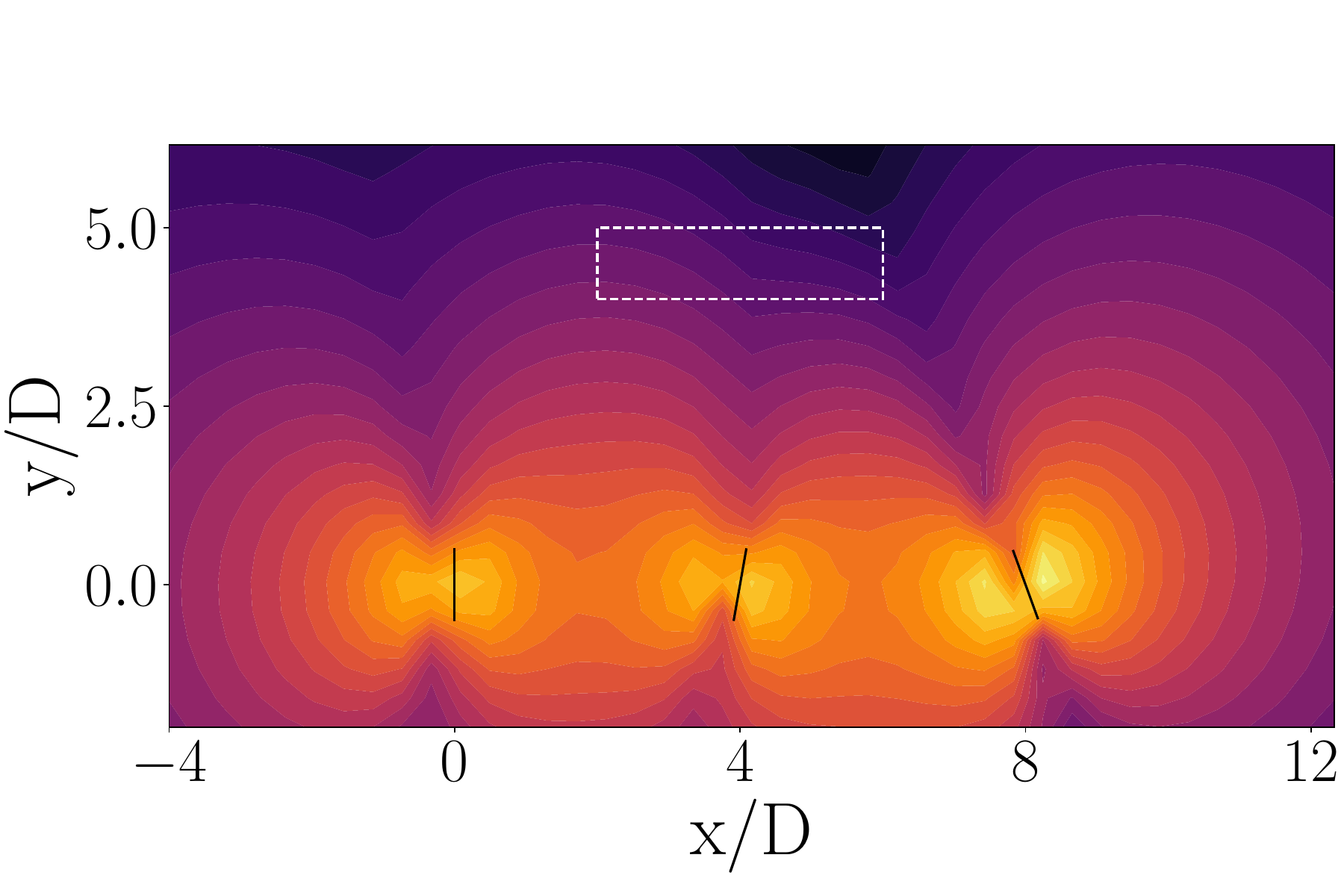} \\
    \rotatebox{90}{\footnotesize\shortstack{\textbf{FAST.Farm}\\\textbf{\& Hagg}}} &
    \includegraphics[width=0.27\textwidth]{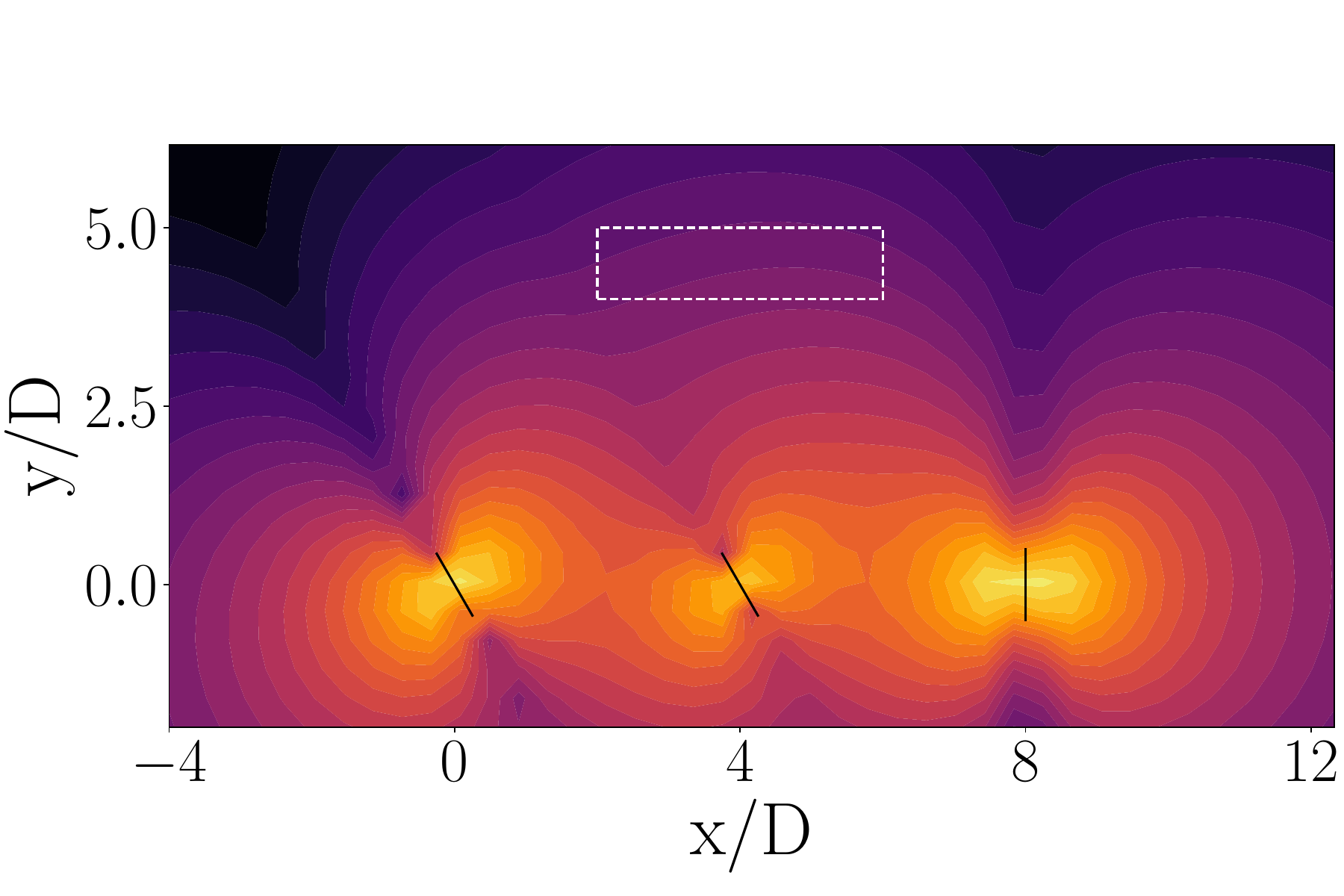} &
    \includegraphics[width=0.27\textwidth]{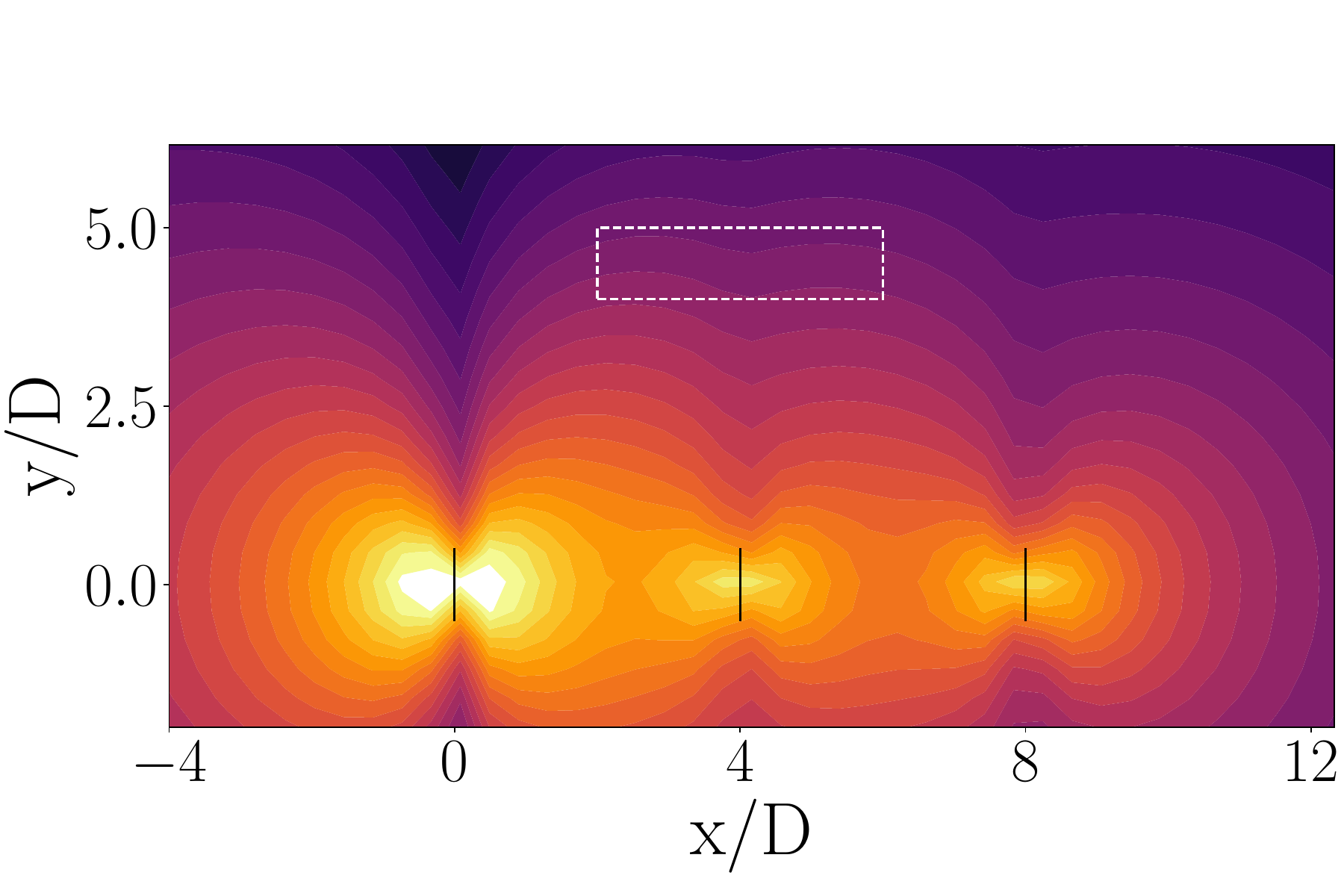} &
    \includegraphics[width=0.27\textwidth]{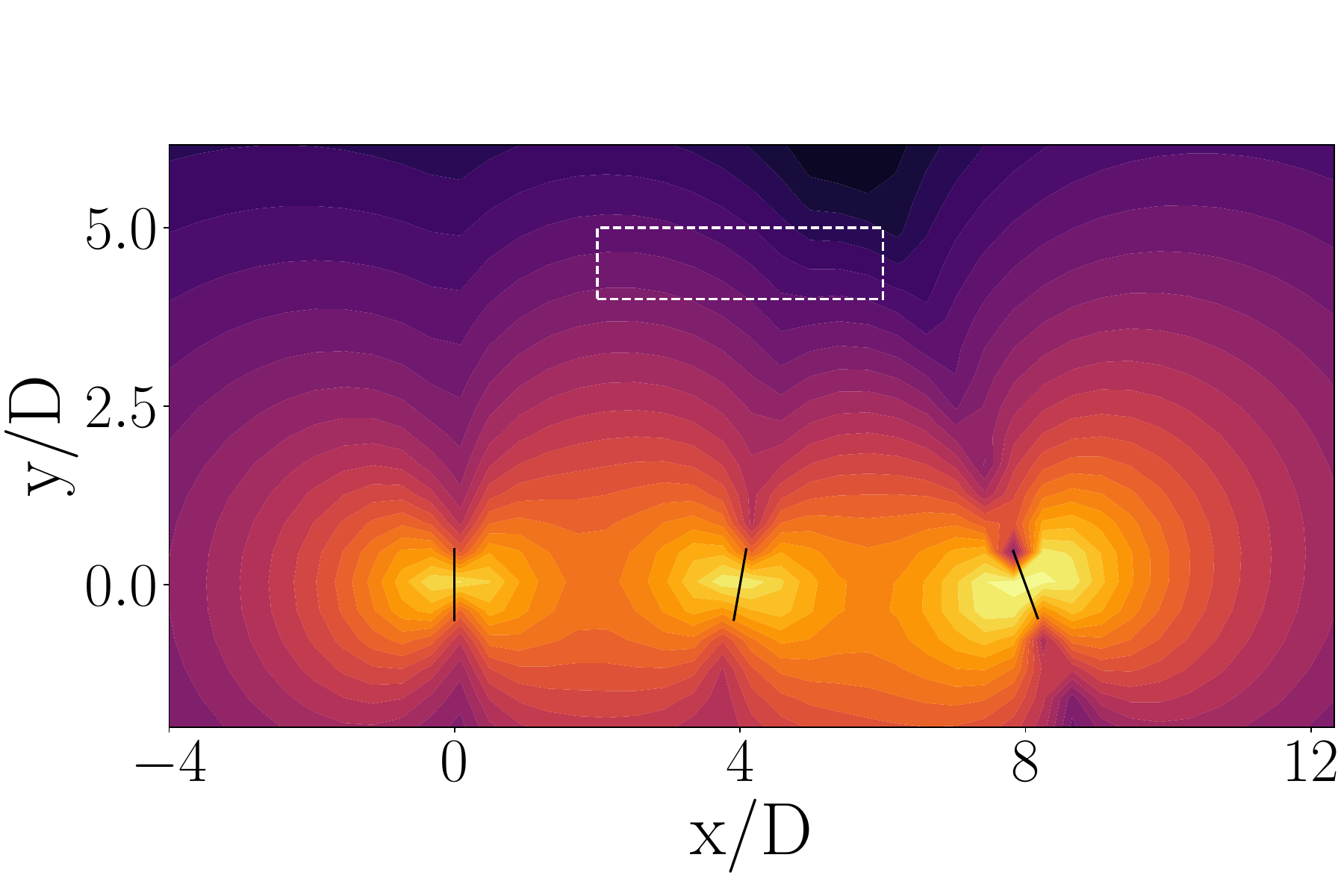} \\
    \rotatebox{90}{\footnotesize\shortstack{\textbf{FLORIS}\\\textbf{\& Hagg}}} &
    \includegraphics[width=0.27\textwidth]{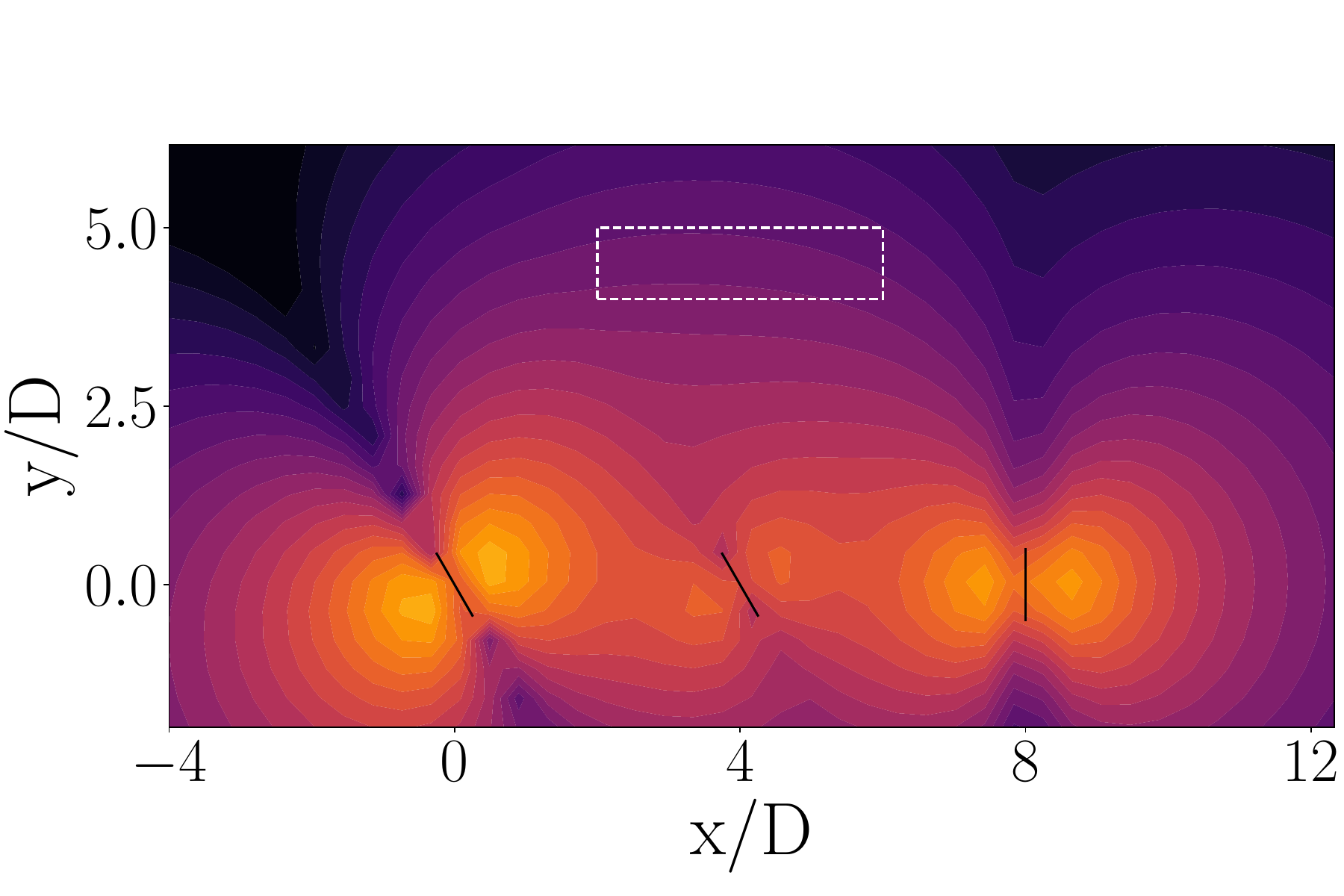} &
    \includegraphics[width=0.27\textwidth]{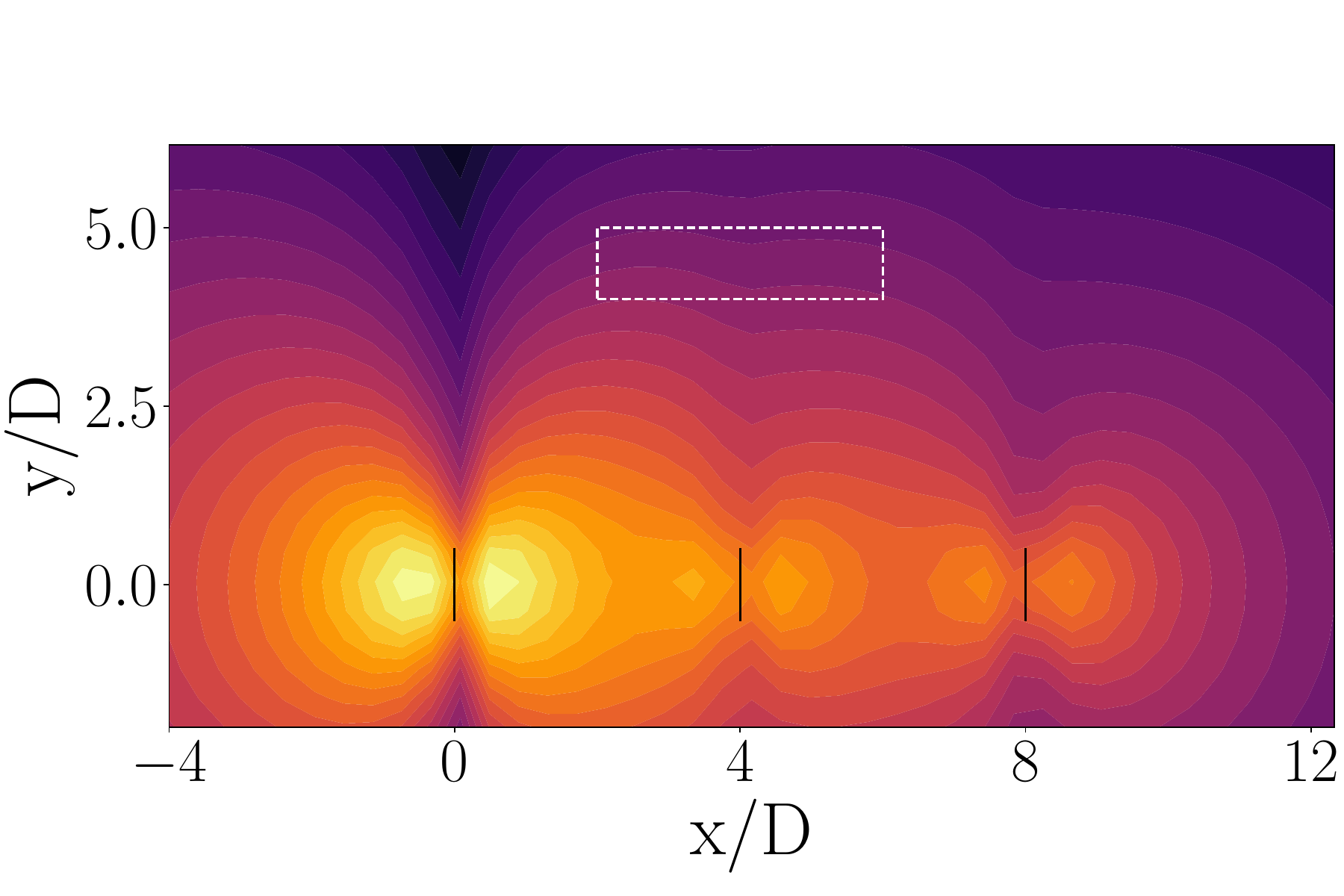} &
    \includegraphics[width=0.27\textwidth]{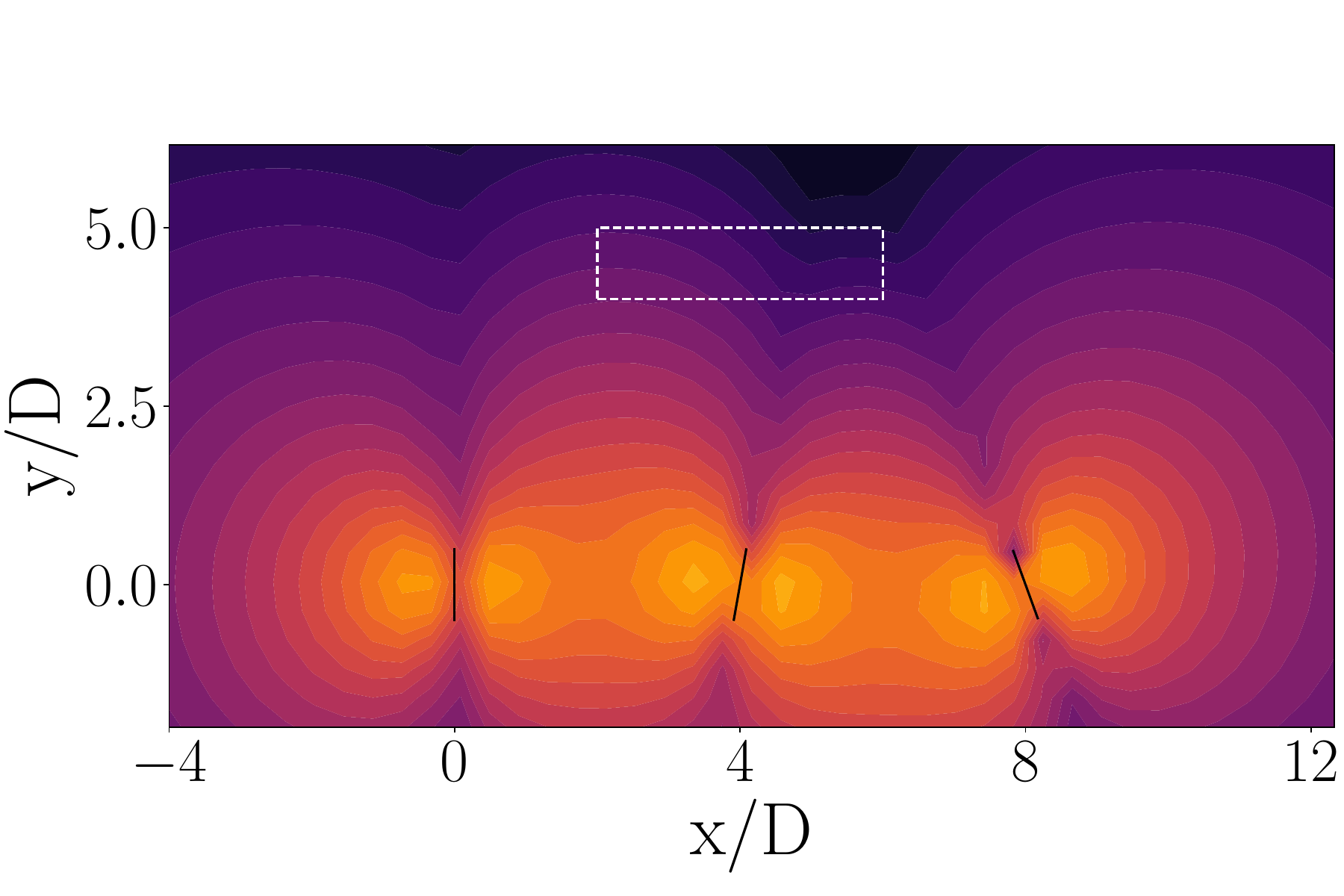} \\
  \end{tabular}%
}
\newlength{\noisetabledepth}
\settodepth{\noisetabledepth}{\usebox{\noisetablebox}}
\newlength{\noisetableheight}
\settoheight{\noisetableheight}{\usebox{\noisetablebox}}
\newlength{\noisetabletotal}
\setlength{\noisetabletotal}{\dimexpr\noisetableheight+\noisetabledepth\relax}
\begin{figure}[htbp!]
    \centering
    \begin{tabular}{cc}
    \hspace{-5mm}
        \raisebox{\noisetabledepth}{\usebox{\noisetablebox}} &
        \includegraphics[height=0.95\noisetabletotal]{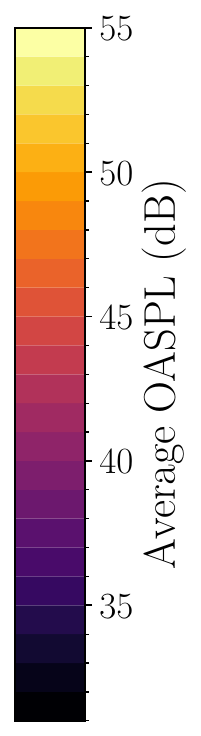}
    \end{tabular}
    \caption{Comparison of three wind farm noise modelling approaches for three operational conditions. The control region is denoted for reference. The OASPL is averaged in time to reduce the amplitude modulation oscillations.}
    \label{fig:Noise_validation}
\end{figure}

\section{\ff and \floris policy comparison}
\label{app:ff_floris}
The \textit{Noise Steering} policy deployed on \ff is analysed over a complete episode to compare the state-action transitions. An episode with a steady free wind speed of $U_\infty=13.2~\ms$ is selected for the comparison. 

\begin{figure}[htbp]
\centering 
     \begin{subfigure}[t]{0.75\textwidth}
        \centering
        \includegraphics[width=\linewidth]{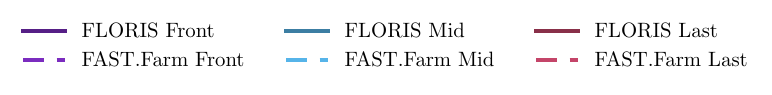}

    \end{subfigure}

    \centering   
     \begin{subfigure}[t]{0.45\textwidth}
        \centering
        \includegraphics[width=\linewidth]{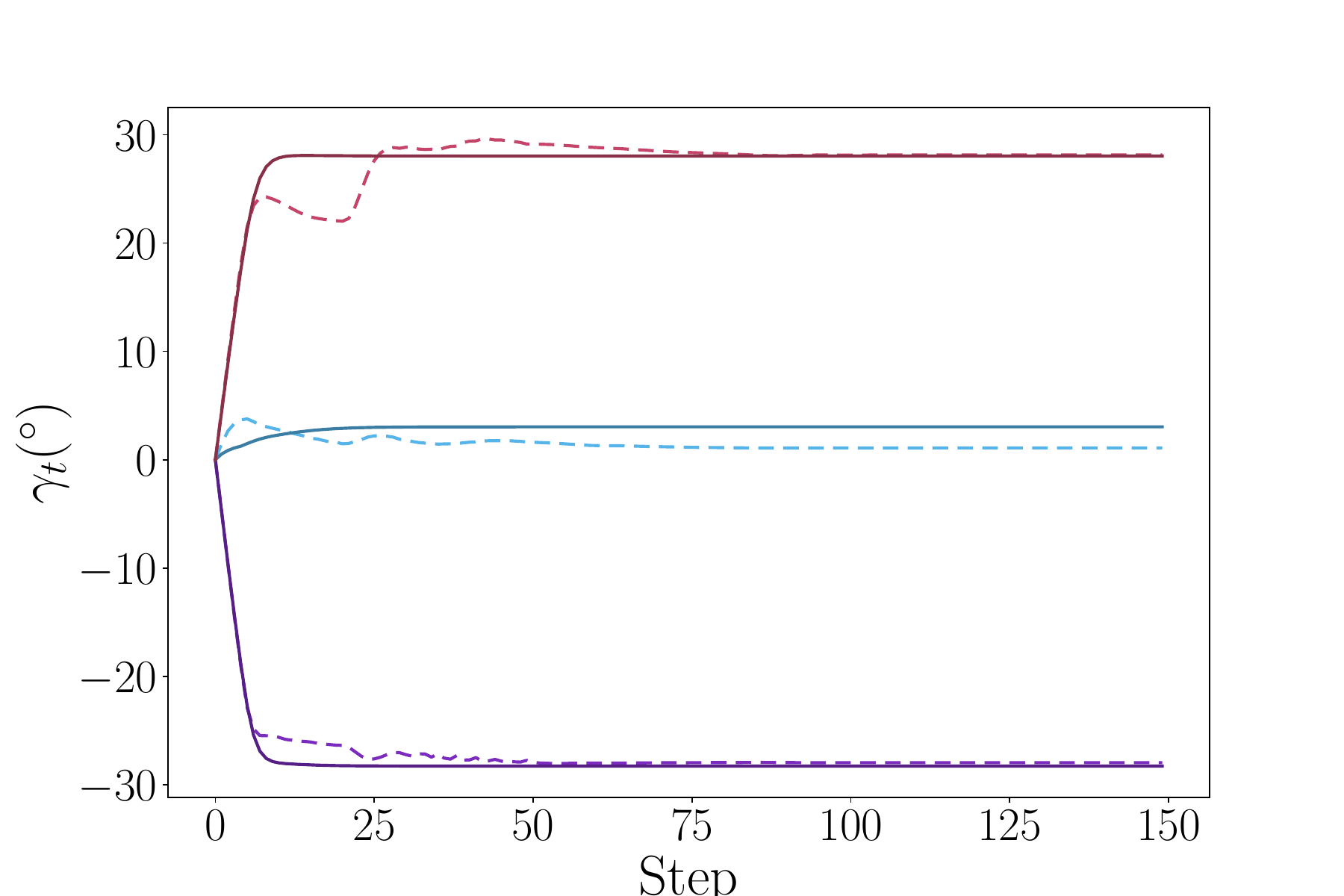} 
        \caption{Nacelle yaw.}
        \label{fig:FF_Floris_episode_yaw}
    \end{subfigure}
    \begin{subfigure}[t]{0.45\textwidth}
        \centering
        \includegraphics[width=\linewidth]{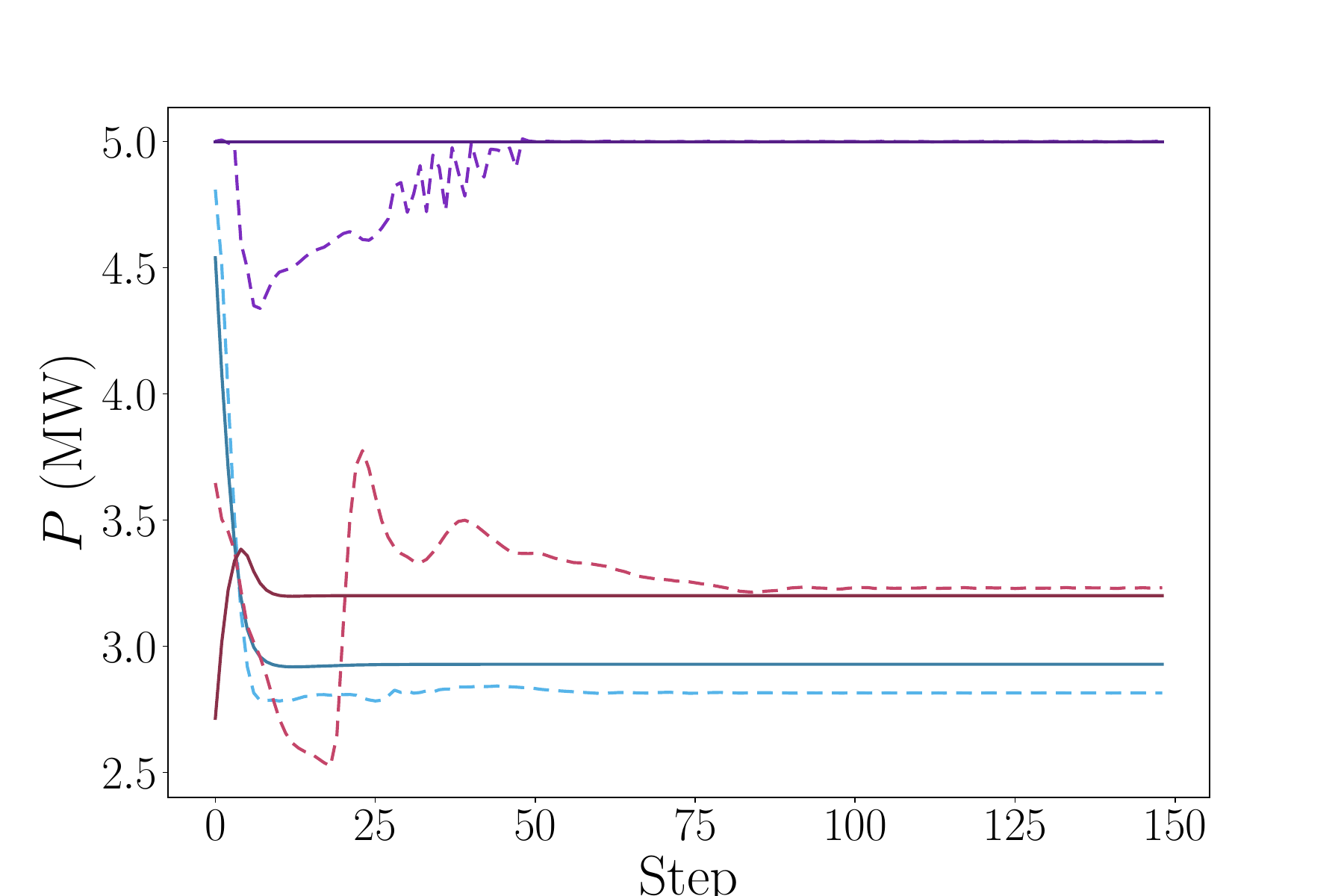} 
        \caption{Wind turbine power.}
        \label{fig:FF_Floris_episode_powers}
    \end{subfigure}
    
    \caption{Turbine-scale episode comparison for the \textit{Noise Steering} policy evaluated in \ff and \floris for $U_\infty=13.2~\ms$. Only the first row of turbines is displayed.}
    \label{fig:FF_Floris_episode_turbines}
\end{figure}

\autoref{fig:FF_Floris_episode_turbines} illustrates the yaw actions and turbine power outputs throughout the episode for both environments. For simplicity, only the first row of turbines is shown. \autoref{fig:FF_Floris_episode_yaw} presents the yaw actions generated by the \textit{Noise Steering} policy. Once the \ff wake dynamics have reached a stable state, the RL policy converges to yaw values close to those obtained in \floris. A similar behaviour is observed for the individual turbine power outputs in \autoref{fig:FF_Floris_episode_powers}.

These results indicate that, despite being trained in a steady-state solver, the RL policy is able to operate under the transient wake dynamics of \ff once the static modelling discrepancy between the two environments has been reduced through calibration.

\begin{figure}[htbp]
    \centering
   \begin{subfigure}[t]{0.45\textwidth}
        \centering
        \includegraphics[width=\linewidth]{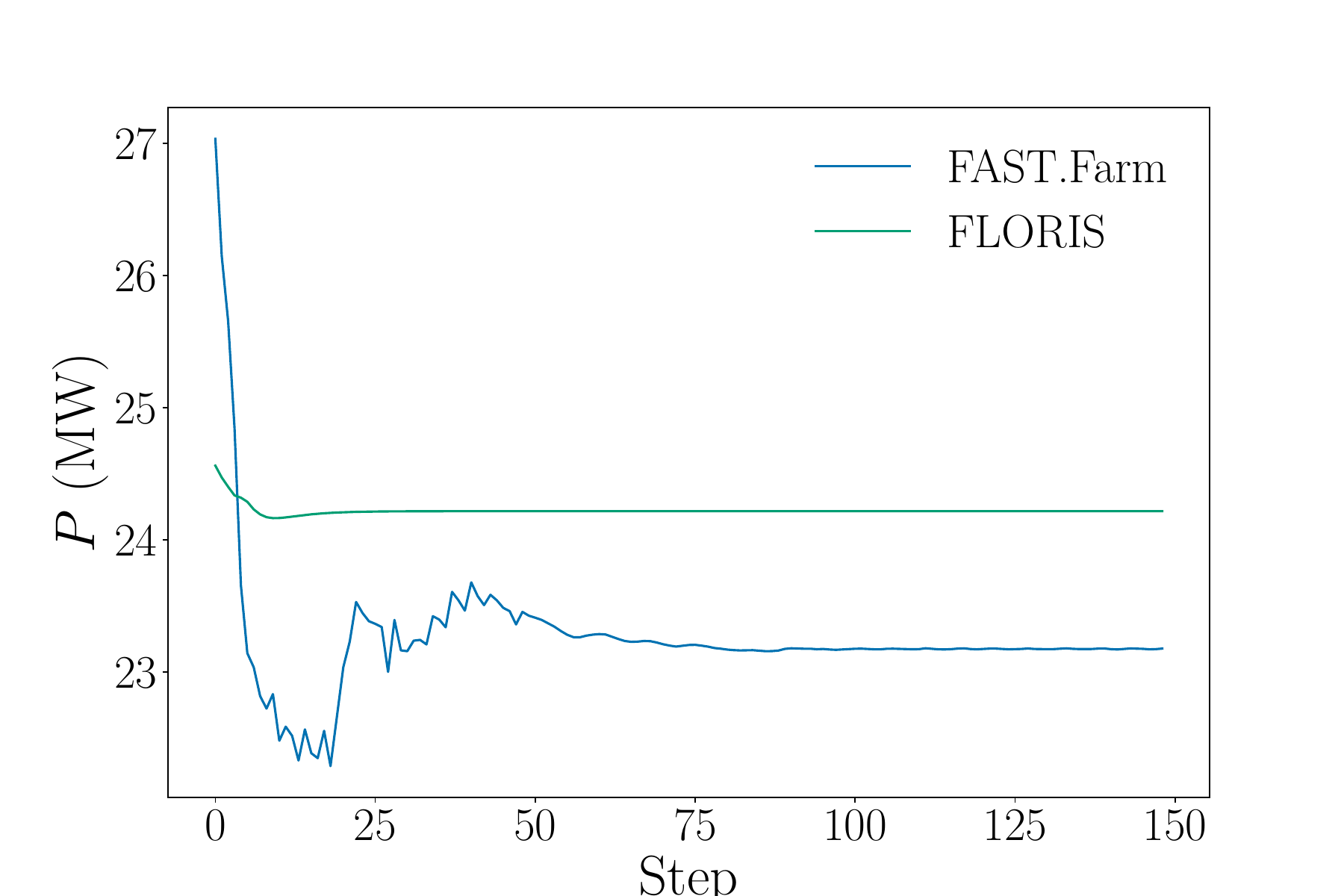} 
        \caption{Farm power.}
        \label{fig:FF_Floris_episode_power}
    \end{subfigure}
    \begin{subfigure}[t]{0.45\textwidth}
        \centering
        \includegraphics[width=\linewidth]{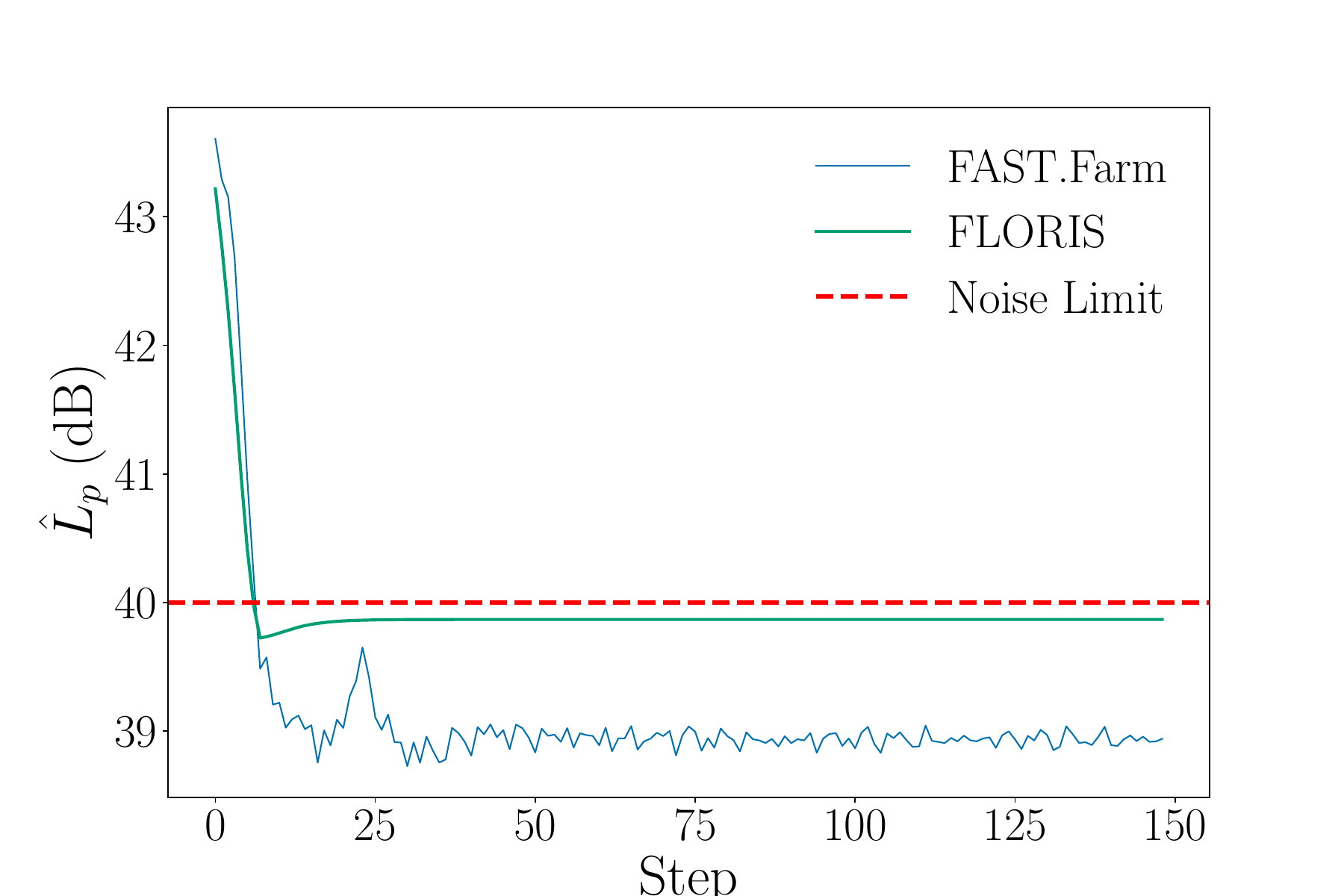} 
        \caption{Acoustic emissions at the control region.}
        \label{fig:FF_Floris_episode_acoustics}
    \end{subfigure}
    \caption{Farm-scale episode comparison for the \textit{Noise Steering} policy evaluated in \ff and \floris. Steady free wind speed of $U_\infty=13.2~\ms.$}
    \label{fig:FF_Floris_episode_Farm}
\end{figure}

Finally, \autoref{fig:FF_Floris_episode_Farm} shows the wind farm power generation and noise emissions throughout the episode. A modelling gap similar to that observed in \autoref{fig:FF_sweep} can be identified. As shown in \autoref{fig:FF_Floris_episode_power}, the \textit{Noise Steering} policy requires approximately 75 steps to stabilize the \ff wake dynamics.

\end{document}